\pdfoutput=1

\documentclass{article}

\usepackage{arxiv}
\usepackage[numbers]{natbib} 
\usepackage[utf8]{inputenc} 
\usepackage[T1]{fontenc} 
\usepackage{hyperref}       
\usepackage{url}            
\usepackage{booktabs}       
\usepackage{amsmath,amsfonts,bm, amssymb}
\usepackage{mathtools}      
\usepackage{nicefrac}       
\usepackage{microtype}      
\usepackage{lipsum}		
\usepackage{graphicx}
\usepackage{float}
\usepackage{subfig}
\usepackage{adjustbox}
\usepackage{doi}
\usepackage{comment}
\usepackage{multicol, multirow}
\usepackage[dvipsnames]{xcolor}
\usepackage{bm}

\usepackage[utf8]{inputenc} 
\usepackage[T1]{fontenc}    
\usepackage{hyperref}       
\usepackage{url}            
\usepackage{booktabs}       
\usepackage{amsfonts}       
\usepackage{nicefrac}       
\usepackage{microtype}      
\usepackage{lipsum}
\usepackage{graphicx}
\graphicspath{ {./images/} }
\usepackage{setspace}

\usepackage{array}
\DeclareMathOperator{\E}{\mathbb{E}}

\usepackage{cleveref}

\title{Temporal Distribution Shift in Real-World Pharmaceutical Data: Implications for Uncertainty Quantification in QSAR Models}

\author{
    \href{https://orcid.org/0000-0002-0047-0964}{\includegraphics[scale=0.06]{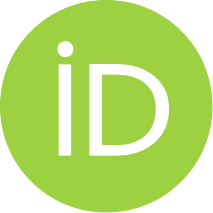}\hspace{1mm}Hannah Rosa Friesacher$^{1,2}$} \And
    \href{https://orcid.org/0000-0001-5598-0286}{\includegraphics[scale=0.06]{figures/orcid.pdf}\hspace{1mm}Emma Svensson$^{2,4}$} \And 
    \href{https://orcid.org/0000-0002-9808-1683}{\includegraphics[scale=0.06]{figures/orcid.pdf}\hspace{1mm}Susanne Winiwarter$^{6}$} \AND
    \href{https://orcid.org/0000-0002-7271-0824}{\includegraphics[scale=0.06]{figures/orcid.pdf}\hspace{1mm}Lewis Mervin$^{3}$} \And 
    \href{https://orcid.org/0000-0002-4901-7650}{\includegraphics[scale=0.06]{figures/orcid.pdf}\hspace{1mm}Adam Arany$^{2}$} \And 
    \href{https://orcid.org/0000-0003-4970-6461}{\includegraphics[scale=0.06]{figures/orcid.pdf}\hspace{1mm}Ola Engkvist$^{1,5}$}
    \AND
    $^1$~Molecular AI, Discovery Sciences\\ 
    AstraZeneca R\&D \\
    Gothenburg, 431 83 Sweden  \\
    \And
    $^2$~ESAT-STADIUS, \\
    KU Leuven, \\
    3000 Belgium  
    \And
    $^3$~Molecular AI, Discovery Sciences \\ 
    AstraZeneca R\&D \\
    Cambridge, CB2 0AA UK \\
    \AND
    $^4$~ELLIS Unit Linz \&\\
    \textbf{Institute for Machine Learning} \\ 
    Johannes Kepler University Linz \\
    Linz, 4040 Austria  
    \And
    $^5$~Department of Computer \\
    \textbf{Science and Engineering} \\ 
    Chalmers University of Technology \\ 
    Gothenburg, 412 96 Sweden
    \AND
    $^6$~ Drug Metabolism and Pharmacokinetics, Research and\\
    \textbf{Early Development Cardiovascular, Renal and Metabolism (CVRM),}\\
    BioPharmaceuticals R\&D, \\ 
    AstraZeneca, Gothenburg, 431 83 Sweden \\     
}
\date{}

\renewcommand{\headeright}{}
\renewcommand{\undertitle}{}

\hypersetup{
pdftitle={Temporal Distribution Shift in Real-World Pharmaceutical Data: Implications for Uncertainty Quantification in QSAR Models.},
pdfsubject={q-bio.NC, q-bio.QM},
pdfauthor={Hannah Rosa Friesacher and Emma Svensson and Susanne Winiwarter and Lewis Mervin and Adam Arany and Ola Engkvist},
pdfkeywords={uncertainty quantification, probability calibration, temporal evaluation, distribution shift, deep learning, drug discovery, molecular property prediction},
}

\begin{document}
\maketitle              

\begin{abstract}
  The estimation of uncertainties associated with predictions from quantitative structure-activity relationship (QSAR) models can accelerate the drug discovery process by identifying promising experiments and allowing an efficient allocation of resources. Several computational tools exist that estimate the predictive uncertainty in machine learning models. However, deviations from the i.i.d. setting have been shown to impair the performance of these uncertainty quantification methods. We use a real-world pharmaceutical dataset to address the pressing need for a comprehensive, large-scale evaluation of uncertainty estimation methods in the context of realistic distribution shifts over time. We investigate the performance of several uncertainty estimation methods, including ensemble-based and Bayesian approaches. Furthermore, we use this real-world setting to systematically assess the distribution shifts in label and descriptor space and their impact on the capability of the uncertainty estimation methods.
  Our study reveals significant shifts over time in both label and descriptor space and a clear connection between the magnitude of the shift and the nature of the assay. Moreover, we show that pronounced distribution shifts impair the performance of popular uncertainty estimation methods used in QSAR models.
  This work highlights the challenges of identifying uncertainty quantification methods that remain reliable under distribution shifts introduced by real-world data. 

\end{abstract}

\newpage
\section{Introduction}
\label{sec:introduction}
The development of new therapeutic agents is a time- and resource-consuming process, characterized by high failure rates and development spans of over a decade until a compound can be put on the market \citep{Singh2023, Laermann2021}.
The use of artificial intelligence (AI), or more precisely, machine learning (ML) approaches, can contribute to easing these problems by using the extensive amount of data produced in the drug discovery pipeline to train computational models that can effectively support future projects with their expert knowledge \citep{Sadybekov2023}.
During early-stage drug discovery, a part of the vast chemical space is screened to identify promising molecular compounds, which are subsequently optimized to achieve the desired properties \citep{Hertzberg2000}. 
The large scale and complexity of this early-stage screening make it an ideal application for ML models with their high computational power and predictive abilities \citep{Sadybekov2023, Bleakley2009}.
Quantitative structure-activity relationship (QSAR) models are well-established in computer-aided drug discovery for identifying compounds with desired features.
They enable the prediction of biological activities or properties of chemical compounds based on their molecular structure.
However, the reliability of these approaches is crucial to optimally support an informed decision-making process, which ultimately saves money and time in the lengthy and costly drug discovery pipeline.

Uncertainty quantification is a powerful tool to increase the reliability of ML models and the confidence in deploying them to real-world applications \citep{Apostolakis1990}. 
Various sources can lead to uncertainty in the predictions obtained from neural networks.
A common classification found in literature is the distinction between aleatoric uncertainty, which originates from uncertainty in the data, and epistemic sources, which quantifies uncertainty inherent in the choice of model \citep{Hullermeier2019, Gruber2023}.
Optimally, estimates of the predictive uncertainty should represent the total uncertainty originating from these different sources. 
Uncertainty quantification methods can be classified into Bayesian approaches \citep{Neal2012, Blundell2015, Izmailov2021, Kim2021}, ensemble-based models \citep{Laksh2017, Gal2016}, conformal predictors \citep{Angelopoulos2021, Taquet2022}, evidential learning \citep{Sensoy2018, Wang2023a, Wang2023b, Oh2022, Soleimany2021} and distance-based approaches \citep{Liu2020}.
Furthermore, multiple techniques exist that can improve the uncertainty estimates post hoc by calibrating them using a simple function trained on a separate calibration dataset \citep{Platt1999, Vovk2014, Zadrozny2002}. 
Many of these computational tools have been explored for drug discovery applications to enable the estimation of predictive uncertainties in molecular property prediction tasks \citep{Mervin2021, Yu2022}.
However, available uncertainty quantification methods vary in their ability to capture all sources of uncertainty correctly, and there is no clear agreement in previous studies on which approach estimates these uncertainties most reliably \citep{Wang2023b, Dheur2023, Schweighofer2023, Mervin2021_prf, Rayka2024, Fan2024, Friesacher2024c, Hirschfeld2020, Mervin2020_va, Dutschmann2023}.

Furthermore, the available uncertainty quantification methods have primarily been evaluated on public data lacking temporal information about the measurements, which is needed to perform data splits that cohere with the history of the assay of interest.
Due to this lack of temporal information, the use of temporal splitting techniques for cross-validation is not possible, which is needed to realistically evaluate model performance over time, as reported by \citet{Sheridan2013} for classification and \citet{Landrum2023} for regression tasks.
Alternative splitting strategies that do not require temporal input include random splits or approaches that are based on the chemical structure of the chemical compounds. 
However, these methods are usually too optimistic or pessimistic compared to the true prospective prediction as they do not reflect the evolution of data in real-world pharmaceutical drug discovery projects \citep{Sheridan2013}.

The first part of this work investigates the evolution of real-world pharmaceutical data and the resulting distribution shift.
\citet{Dundar2007} reported an intrinsic assumption of many training algorithms that the data is independent and identically distributed (i.i.d).
While this assumption is foundational for traditional ML models, it imposes significant constraints and oversimplifies the complexities of realistic scenarios \citep{Cao2022}. 
Consequently, the simplified problems may fail to accurately represent or address the challenges inherent in real-world datasets, such as the pharmaceutical data included in this study.
In the context of probability calibration, deviations from the i.i.d. setting have been shown to impair the performance of common uncertainty estimation methods previously reported to improve model calibration under i.i.d. conditions \citep{Ovadia2019, Koh2021}.

Therefore, the second part of this work compares common uncertainty estimation approaches that employ real-world temporal splits to evaluate model performance in a more realistic setting.
In binary classification problems, neural networks typically give probability-like predictions that can be directly interpreted as an estimate of the confidence in the prediction. 
Previous work has concluded that modern neural networks often fail to give realistic estimates of the uncertainty associated with a prediction in classification tasks, resulting in poorly calibrated models \citep{Mervin2021, Yu2022, Guo2017}. 
Several approaches exist in the literature that use more sophisticated techniques to improve the reliability of these uncertainty estimates.
For a more straightforward comparison between the uncertainty estimation methods used in this work, we classify them into two categories, namely train-time uncertainty estimation approaches and post hoc probability calibration methods.

Train-time uncertainty quantification approaches refer to Bayesian methods or ensemble-based techniques inspired by the Bayesian framework to estimate the posterior distribution of predictions from a set of models \citep{Gal2016, Laksh2017, Sheridan2012}.
These approaches include uncertainty by accounting for model variance, which increases when the neural network is overfitting, or the test instance lies outside the domain of the training data. 
We consider three methods for train-time uncertainty estimation in this work.
Deep ensembles and Monte Carlo (MC) dropout aim to improve the performance by obtaining numerous base estimators to determine the model variance \citep{Hullermeier2019, Laksh2017, Gal2016}.
Furthermore, we compare the ensemble-based strategies with a full Bayesian neural network trained with the Bayes-by-Backprop approach \citep{Blundell2015}.
The posterior distribution accounts for model variance in the Bayesian setting, where the network's parameters are treated as random variables rather than point estimates and which therefore provides a natural solution for including epistemic uncertainty. 
Bayes-by-Backprop allows to quickly obtain samples from the posterior distribution of the neural network weights by using a variational approximation scheme.

While these train-time uncertainty quantification approaches aim to achieve better uncertainty estimation by accounting for the epistemic uncertainty,
post hoc calibration approaches improve model calibration by applying an additional post-processing step to the scores retrieved from a separately trained classifier.
These methods require a separate dataset, called a calibration set, to train the calibrating function used in the post-processing step.
For this study, we tested two post hoc calibration techniques, including the commonly used Platt scaling approach \citep{Platt1999} and Venn-ABERS predictors \citep{Vovk2014}, which were previously shown to enhance the probability calibration of classifier predictions \citep{Mervin2020_va, Niculescu2005}.
Platt scaling fits a logistic regression to the classification scores of the calibration set to counteract over- or underfitted uncertainty estimations, while Venn-ABERS predictors use the more flexible isotonic regression functions to calibrate the probability point estimates.

To our knowledge, only a few studies have addressed the performance of uncertainty estimation approaches under temporal shifts.
In some of these works, temporal splitting approaches are applied to ChEMBL \citep{Zdrazil2024} data, using the publication date as a reference \citep{Wang2023b}.
As the date of publication does not correspond to the date when the experiment was conducted, it remains questionable how accurately this information can reflect the timeline in a pharmaceutical company.
Other studies \citep{Rodriguez2022, Stoyanova2023} used internal data from pharmaceutical companies with the necessary information to perform proper temporal splits.
\citet{Rodriguez2022} studied the performance of multitask graph neural networks for uncertainty estimation focusing on intrinsic clearance data.
Another recent work compares the uncertainty estimation of various regression models for pharmacokinetic property prediction of potential drug molecules \citep{Stoyanova2023}.
However, both of these studies use training data from a fixed time span for all experiments and, therefore, do not address model performance over time.
Furthermore, they do not address shifts in the data caused by the temporal splitting strategy.
\citet{Svensson2024arxiv} recently published an extensive temporal study comparing uncertainty estimation methods trained on drug-target interaction data with and without censored data.
While this study provides a comprehensive guide on handling uncertainty estimation methods for regression tasks, a comparable large-scale study that applies temporal splitting strategies to different biological assays has yet to be published for classification approaches.

In this work, we aim to address these gaps by assessing the performance of different uncertainty estimation approaches using single-task models over time and in the context of assay-specific distribution shifts in the data.
The models were trained on an internal dataset, which has already been studied in previous works \citep{Svensson2024arxiv, Friesacher2024a, Friesacher2024b}.
This dataset includes drug-target interaction data from different biochemical assays, providing the additional information required to perform temporal splits.
First, we analyze the history of the individual assays and how the data distribution shifts in label and descriptor space over time. 
Next, we compare available probability calibration and train-time uncertainty quantification methods and explore possible connections between their performance and data shifts.

\section{Methods}
\label{sec:methods}
The following section is structured into three parts to examine the material and methods used in this study.
The first section describes the assay data and the splitting strategy used to generate splits representing different time spans in the assay history.
The second part addresses the modeling approaches, including the base estimators and the more sophisticated uncertainty estimation approaches.
Finally, the last section provides insight into the experiments and metrics used to compare the uncertainty estimation approaches comprehensively.

\begin{figure}[h!]
        \centering
        \includegraphics[width=\linewidth]{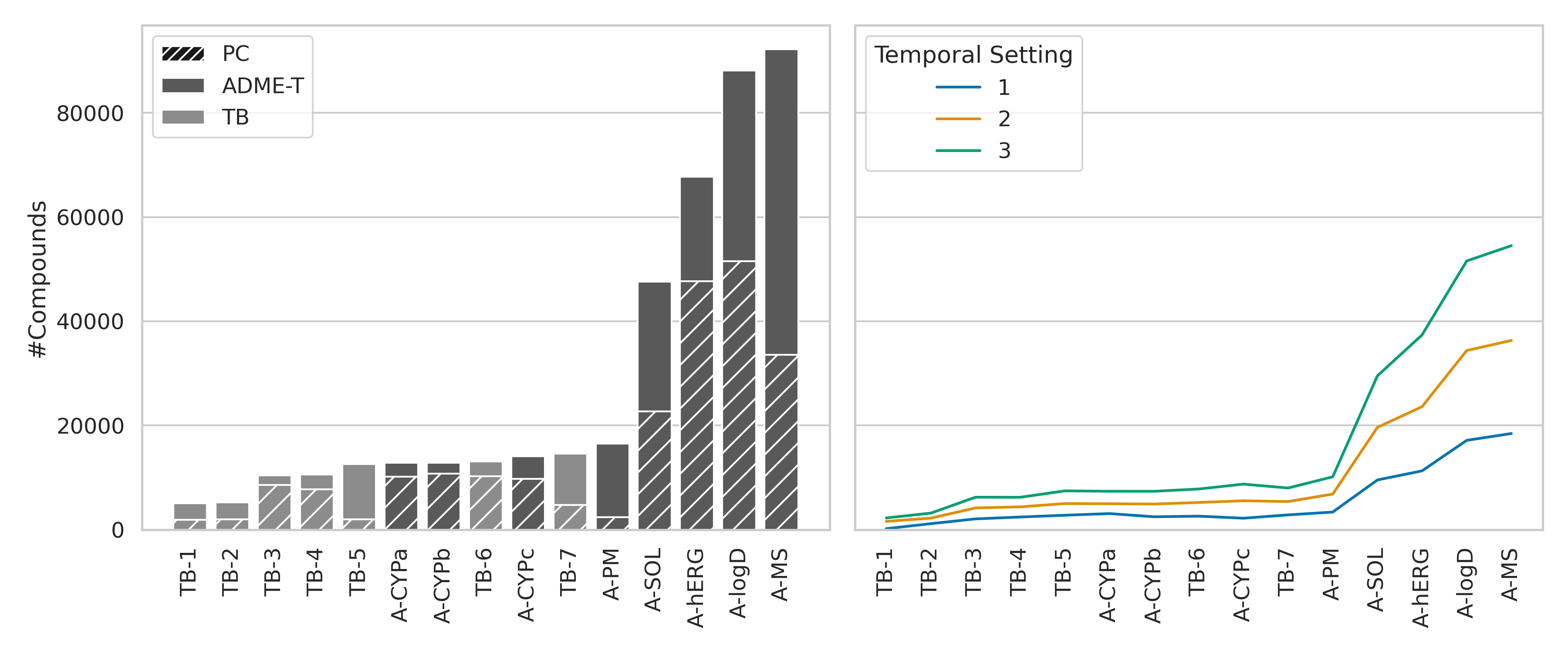}
        \caption{\textbf{Overview of dataset sizes.} The left panel plots the size of the individual assays ordered according to assay size. The striped areas in the bar indicate the amount of compounds belonging to the preferred class (PC) in each assay.
        The right panel shows the amount of training data in each temporal setting across all assays, with 1, 2, or 3 time spans used for training.}
        \label{fig:data}
    \end{figure}

\subsection{Data}
\label{sec:data}
    
This study uses internal data from 15 biological assays to gain insight into the properties of real-world pharmaceutical data and subsequently train binary classifiers for each assay separately. 
Parts of the dataset \citep{Friesacher2024a, Friesacher2024b} or the whole dataset \citep{Svensson2024arxiv} have already been used in previous studies.
The included assays are diverse and represent different optimization problems typically addressed during the drug discovery workflow to enhance the pharmacokinetic and pharmacodynamic properties of a drug candidate.
Furthermore, the assays exhibited different sizes, modeled endpoints, and ratios of the preferred class in the individual datasets.
The total size of the assays is illustrated in the left panel of Figure \ref{fig:data}. 
In addition, Table \ref{tab:data} provides detailed information on the assays used in this study. 
The assays were assigned to two categories, Target-Based (TB) and ADME-T, and subsequently labeled based on size and category affiliation.
The labels of the ADME-T assays were created from the class name and the assay description together with the

\begin{table*}[h!]
\caption{\textbf{Overview of the assay data.} Details of the assays included in this study, including assay size, assay unit, and modeled endpoint. Descriptions of the ADME-T assays are shown. The ratio of compounds belonging to the preferred class (PC) is reported for each assay. The last column indicates the corresponding threshold $T$ used to assign compounds to classes and if the PC lies above or below $T$ (PC $ </> T$)}.
    \centering
    \begin{tabular}{lrccccc}   
    \toprule
          & Assay & Assay & Ratio & Assay & Modeled & Threshold $T$\\
         Abbreviation & Description & Size & PC & Unit & End-point &  (PC $ </> T$) \\
        \midrule
        \textbf{Target-Based}\\
            TB-1 & NA & 5,082 & 0.38 & µM & pIC50 &  $>$ 6 \\
            TB-2 & NA & 5,237 & 0.39 & µM & pIC50 & $>$ 6 \\
            TB-3 & NA & 10,465 & 0.82 & µM & pIC50 &  $>$ 6 \\
            TB-4 & NA & 10,624 & 0.73 & µM & pIC50 &  $>$ 6 \\
            TB-5 & NA & 12,612 & 0.16 & µM & pEC50 &  $>$ 6 \\
            TB-6 & NA & 13,093 & 0.79 & µM & pIC50 &  $>$ 6 \\
            TB-7 & NA & 14,605 & 0.33 & µM & pIC50 &  $>$ 6 \\
        \midrule
        \textbf{ADME-T}\\
            A-CYPa & CYP3A4 & 12,875 & 0.79 & µM & pIC50 &  $<$ 5 \\ 
            A-CYPb & CYP2C9 (I) & 12,876 & 0.84 & µM & pIC50 & $<$ 5 \\
            A-CYPc & CYP2C9 (II) & 14,062 & 0.30 & µM & pIC50 &  $<$ 5 \\               
            A-PM & Permeability & 16,511 & 0.15 & 1e-6cm/s & logP &  $>$ 1 \\
            A-SOL & Solubility & 47,607 & 0.48 & µM & logS &  $>$ 2 \\ 
            A-hERG & Toxicity & 67,687 & 0.70 & µM & pIC50 &  $<$ 5 \\
            A-logD & Lipophilicity & 88,114 & 0.58 & - & logD &  $>$ 3 \\
            A-MS & Metabolic Stability & 92,161 & 0.36 & µl/min/1e6 & logMS &  $<$ 1 \\  
        \bottomrule
    \end{tabular}
    \label{tab:data}
\end{table*}

 prefix ``A`` to indicate theaffiliation to the ADME-T category (e.g., A-logD for the lipophilicity assay). 
The TB assays were ordered according to size and numbered consecutively (TB-1 for the smallest to TB-7 for the largest assay).    

\paragraph{TB Assays.}

The TB category includes project-specific assays from activity screens to identify active substances on a specific target of interest.
Active substances are compounds that modulate the function of a protein, for example, by inhibiting or activating the target. 
This work includes seven TB assays, with assay sizes ranging from 5,082 to 14,605 measured compounds.
As opposed to the ADME-T assays, further specifics regarding these biological assays cannot be disclosed due the proprietary constraints.

\paragraph{ADME-T Assays.}

Assays in the ADME-T category typically assess the pharmacokinetic properties and toxicity profile of a drug candidate. 
These properties are connected to the absorption, distribution, metabolism, and excretion (ADME) of a compound, while the toxicity screens identify compounds that hit unintended targets.
The ADME-T category comprises assays that assess the general features of a compound, which are typically relevant for the success of a promising compound in the drug discovery and development pipeline \citep{Dimasi2001, Van2003}. 
The assays, which include data from various projects, are usually comparatively large.
In our study, eight ADME-T assays were used, including five large assays with measurement numbers between 16,511 and 92,161 and three smaller assays comprising 12,875 and 14,062 data points, which measure interactions with Cytochrome P450 (CYP).

As opposed to the TB assays, the ADME-T assays included in this study are widely used in the drug discovery process, which allows the disclosure of more detailed descriptions of the assays.
The CYP assays measure the inhibition of one of the two CYP isoforms, CYP3A4 (A-CYPa) and CYP2C9 (A-CYPb and A-CYPc). 
These isoforms play an essential role in drug metabolism and the detection of drug-drug interactions \citep{Deodhar2020, Ioannides1996, Furge2006}.
Two distinct assay types are available, exploring different types of interactions with the CYP isoforms.  
The CYP2C9 (I) and CYP3A4 assays measure drug molecule disappearance using liquid chromatography-mass spectrometry, while the CYP2C9 (II) assay measures CYP inhibition using a fluorescent substrate.
In both assays, weaker interactions with the CYP protein are usually favorable to avoid rapid decomposition of the drug molecule and drug-drug interactions. 
The permeability assay (A-PM) evaluates the flux of a compound across a Caco-2 cell, reflecting its potential in vivo absorption, which is measured in 1e-6 cm/s \citep{Shah2006}.
High velocities are favorable, indicating a compound's ability to cross biological membranes.
The solubility assay (A-SOL) assesses the maximum concentration of a compound in an aqueous solution at pH 7.4.
A Dimethyl sulfoxide (DMSO) stock solution is used, and the organic solvent is evaporated to obtain a solid sample.
Compounds with high solubility are preferred to allow sufficient dissolution in biological fluids \citep{Di2012}.  
The hERG assay (A-hERG) provides vital insight into a compound's toxicity profile by measuring its inhibiting effects on the human Ether-a-go-go Related Gene (hERG) potassium channel.
Inhibition of hERG is correlated to severe cardiac side effects by prolonging the QT interval \citep{Keating1996}.
Therefore, inhibiting interactions with hERG is usually undesirable.
The lipophilicity of a compound is obtained in the A-logD assay by measuring the logarithm of the distribution coefficient between octanol and aqueous phase at pH 7.4. 
Lipophilicity is crucial since it significantly affects drug absorption, metabolism, and safety.
A logD greater than 3 has previously been identified as a trigger for safety concerns \citep{Waring2010, Chen2013}.
Finally, the metabolic stability assay (A-MS) measures how fast a compound is metabolized in rat hepatocytes.
The in vivo hepatic clearance is measured in µl/min/million cells.
In general, low values for hepatic clearance are desirable, as they imply slower decomposition and, therefore, higher bioavailability of the drug molecule \citep{Masimirembwa2001, di2003}.  

\paragraph{Binary Classification of Compounds.}
The measured values were converted to a logarithmic scale, and a suitable threshold was determined for each assay individually.
Assay-specific thresholds were defined to determine if a compound belongs to the preferred class.
All TB assays,  obtain a compound's inhibiting potency by measuring the IC50, except TB-5, in which the EC50 was used.
The IC50 value measures the compound concentration needed to inhibit half of a protein's activity, while the EC50 value indicates the compound's concentration that triggers half of the maximum possible effect.
Subsequently, these values were converted to pIC50/pEC50 by taking the logarithm of the measurements converted from micromolar ($\mu M$) to molar. 
The preferred classes in the TB assays comprise compounds with a pIC50/pEC50 value above 6, indicating that a substance achieves the desired effect when its IC50/EC50 value is below 1 $\mu M$.
In addition, four ADME-T assays, including the three CYP and the hERG assays, contain pIC50 values.
Since these assays aim to detect interaction with off-targets, lower pIC50 thresholds of 5 were selected to decrease the risk of false negatives.
The preferred class includes compounds with pIC50 values below this threshold.
For A-logD, A-SOL, A-PM, and A-MS, individual thresholds were chosen as shown in Table \ref{tab:data} to assign compounds with desirable properties to the preferred class.

\paragraph{Temporal Split.}

\begin{figure*}[t]
    \vskip 0.2in
    \centering   \includegraphics[width=\textwidth]{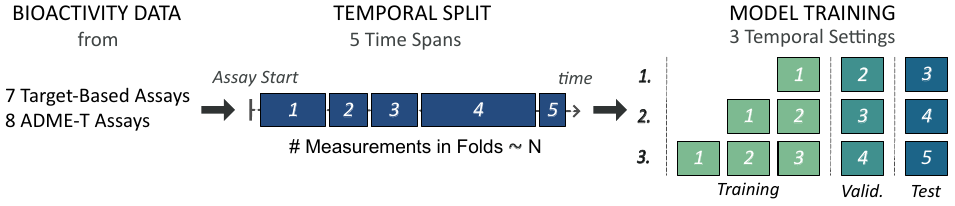}
    \caption{\textbf{Overview of the temporal split and model training.} The data in each assay was assigned to 5 time spans to create three temporal settings, each with increasing amounts of training (Training) data. The subsequent two folds were used for validation (Valid.) and testing (Test). The validation data also served as a calibration set used in post hoc calibration approaches.}
    \label{fig:Split}
    \vskip -0.2in
\end{figure*}

For each assay, we split the data into five roughly equally sized folds using the date of each measurement. 
Each fold represents a specific time span in the history of the assay. 
These folds were then used to set up three experimental settings, using one, two, or three folds for training the QSAR models. 
In each case, the first subsequent fold was used for validation, including model selection and calibration where applicable. 
We only evaluated each setting on the first fold following the validation set for consistency between test sets. 
However, all remaining folds could, in principle, be used.
Figure \ref{fig:Split} illustrates the temporal splitting strategy. 
Considering all assays and settings, 45 separate training datasets were used throughout this work. 
For experiments in which the results of all three settings are used, the assays are labeled with the Assay Abbreviation [Temporal Setting]. 
Naturally, the size of the training sets varies among the temporal settings as shown in the right panel of Figure \ref{fig:data}.

\subsection{Models}
\label{sec:models}

\begin{figure}[h!]
        \centering
        \includegraphics[width=\linewidth]{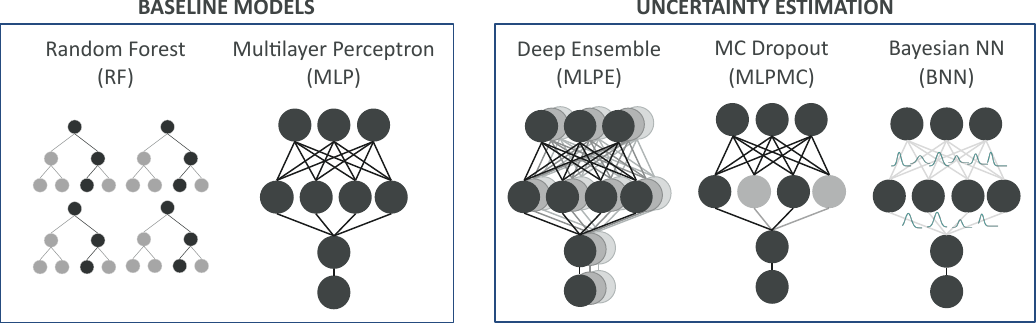}
        \caption{\textbf{Overview of the classification models.} The architectures of the baseline models and train-time uncertainty quantification methods compared in this study are shown. All models were trained in a single-task manner. The hyperparameters of the baselines, RF and MLP, were tuned in an extensive grid search. The baseline MLP was used as the basis for the three uncertainty quantification methods, deep ensembles, MC dropout, and a Bayesian neural network.
        }
        \label{fig:models}
    \end{figure}

Figure \ref{fig:models} provides an overview of the models compared in this study.
All architectures used in this work stem from a Random Forest (RF) or a multilayer perceptron (MLP). 
Both approaches are commonly used in research addressing uncertainty estimation in QSAR modeling \citep{Mervin2020_va, Dutschmann2023, Mervin2021_prf}. 
Furthermore, both approaches can be easily combined with the ECFP fingerprint representation. 
Note that more sophisticated options exist for molecular representations and model architectures, like graph neural networks for molecular graph representations or language models for SMILES representations. 
However, since our study aims to gain insight into uncertainty estimation in QSAR models rather than finding the best approach, or comparing molecular representations, we opted for the simple ECFP representation.
The RDKit package \citep{landrum2006rdkit} was used to generate ECFP fingerprints of length 4096 from the SMILES \citep{Weininger1988smiles} of the compound structures. 
Due to additional computational constraints, we concentrated on RF and MLP models as suitable choices for examining uncertainty quantification in a temporal context.

\paragraph{Model Generation.}
A Python package is publicly available at \url{https://github.com/MolecularAI/uq4dd}, which contains the code used for model generation and evaluation inspired by the design pattern proposed by \citet{hartog2024registries}.
The hyperparameter tuning for the two base estimators, RF and MLP, was performed using an exhaustive grid search.
The binary cross-entropy (BCE) loss was calculated to compare the model performance on a validation set. 
The exact parameter space search is described
in the appendix (Table \ref{tab_supp:hyperparameters}).  
The RF models were generated using scikit-learn \citep{scikit-learn}.
During hyperparameter tuning, the maximum depth of the trees and the required number of estimators of each assay and temporal setting were individually tuned using the validation BCE loss. 
Probability-like outputs were generated from the ratio of decision trees in the RF that classified a test instance as active. 
The MLP models were trained using PyTorch \citep{Paszke2019} with the BCE loss function. 
Similarly, the model selection, including early stopping, was optimized using the validation loss for every assay and temporal setting. 
The network architecture was optimized for the number of hidden units, number of hidden layers, and dropout rate. 
Additionally, the learning rate and scaling factor of a ReduceOnPlateu learning rate scheduler were also optimized. 
Adam was used as an optimization algorithm  \citep{Kingma2014} to train the neural networks.
Probability-like scores were obtained by applying a sigmoid function to the output of the MLP. 

The base estimators were further modulated to generate more sophisticated uncertainty estimation methods.
Train-time uncertainty quantification methods were trained using the MLP base estimators.
Furthermore, post hoc probability calibration methods were applied to selected models.
A detailed description of the train-time and post hoc calibration uncertainty estimation methods can be found below.

\paragraph{Train-time Uncertainty Quantification.}
Train-time uncertainty quantification approaches aim to estimate uncertainty during model training by accounting for uncertainty in the neural network.
They account for model variance, which increases when the model is overfitting, or the test instance lies outside the domain of the training data.
In contrast to post hoc calibration methods, they do not apply a post-processing step to the scores of the classifier. 
In this work, we compare two ensemble-based techniques inspired by the Bayesian theorem: deep ensembles (MLPE) and Monte Carlo (MC) dropout (MLPMC). 
We also include one full Bayesian neural network trained with the Bayes-by-Backprop approach (BNN).
These methods aim to estimate the posterior distribution over the parameters of the neural network\citep{Laksh2017, Gal2016, Sheridan2012}.
Theoretically, the posterior distribution $P(\Theta|D)$ over model parameters $\Theta$, given data $D$, can be computed using the Bayesian theorem

\begin{equation}
P(\Theta|D) = \frac{P(D|\Theta)P(\Theta)}{\int P(D|\Theta)P(\Theta) \,d\Theta\ }.
\end{equation}

When working with high-dimensional posteriors, the calculation of the closed-form solution of the Bayesian equation is usually infeasible due to the intractability of the evidence term in the denominator, which requires solving a highly complex integral.
To circumvent this problem, sampling-based methods are often used that retrieve samples $\Theta := \{\theta_{1},\theta_{2}, ... ,\theta_{N}\}$ from the posterior distribution, so that $\theta_{n} \sim P(\Theta|D)$.
During inference, the predictions of the sampled models are averaged to obtain a mean estimate of the target label $y$ given the descriptor $\bm{x}$:

\begin{equation}
P(y|x, D) \approx \frac{1}{N} \sum_{n = 1}^{N} P(y|x,\theta_{n}).
\end{equation}

Both ensemble-based approaches, deep ensembles (MLPE) and MC dropout (MLPMC), approximate the Bayesian treatment by estimating the predictive uncertainty using numerous base estimators.
Deep ensembles use multiple randomly initialized models as base estimators, corresponding to different local minima in the loss landscape \citep{Laksh2017}.
In this work, 25 base estimators were trained, and their predictions were averaged to obtain a point estimate.
MC dropout applies dropout during inference by setting a number of randomly selected neurons to zero to introduce stochasticity \citep{Gal2016}.
To generate MC dropout (MLPMC) models, 400 forward passes using dropout were aggregated, with the average being the final prediction of the models.

To compare the ensemble-based methods with a full Bayesian approach, we include Bayesian neural networks (BNN) trained with the Bayes-by-Backprop method in the comparison study.
We used a previously published repository for the Bayes-by-Backprop method accessible at \url{https://github.com/ThirstyScholar/bayes-by-backprop} as a template for our implementation of the BNN approach.
In the Bayesian setting, neural network weights are treated as random variables rather than point estimates, which allows model variance to be accounted for in the posterior distribution of the weights.
Since the parameter space is usually high-dimensional, the closed-form solution of the posterior distribution cannot be solved.
Bayes-by-Backprop provides a quick solution for obtaining samples from the approximate posterior distribution of neural network weights $W$ using a variational approximation scheme.
The underlying idea of Bayes-by-Backprop is to learn the optimal parameters $\Theta^*$ of a surrogate distribution that minimizes the Kullback-Leibler (KL) divergence \cite{Kullback1951} between the simpler surrogate $q(W|\Theta)$ and the complex posterior distribution $P(W|D)$

\begin{equation}
 \Theta ^* = \underset{\Theta}{\mathrm{argmin}}\,  KL[q(W|\Theta)|P(W|D)].    
\end{equation}

The calculation of the resulting KL divergence requires the incomputable closed-form solution of the posterior.
Therefore, the evidence lower bound (ELBO) is used to derive a computable loss function that can be used in the backpropagation framework:

\begin{equation}
\mathcal{L}(\Theta, D) = \underset{\Theta}{\mathrm{argmin}}\, KL[q(W|\Theta)|P(W)] - \E_{q(W|\Theta)}(\mathrm{log} P(D|\Theta)). 
\end{equation}

When sampling from the surrogate distribution, the introduced stochasticity prevents using a backpropagation scheme.
To allow the computation of gradients, the local reparametrization trick is applied.
Instead of sampling directly from the proposal function, a deterministic transformation function with learnable parameters is used to convert a sample of parameter-free noise into a sample of the proposal function.
We refer to \citet{Blundell2015} for more technical details of the Bayes-by-Backprop approach.

\paragraph{Post hoc Probability Calibration.}
Two post hoc probability calibration techniques were fitted to each model using the validation set.
These approaches included Platt scaling \citep{Platt1999} and Venn-ABERS (VA) predictors \citep{Vovk2014}. 
Platt scaling fits a logistic regression to the classification scores to counteract over- or underfitted uncertainty estimations \citep{Platt1999}. 
Two isotonic regression functions were trained on the validation set and a given test instance \citep{Vovk2014} for calibration with VA predictors. 
The two isotonic regression functions represent the hypothesis that the test instance is active versus inactive. 
As such, the probabilities obtained from the isotonic regression functions correspond to a lower and an upper bound on the estimated probability. Finally, these bounds were condensed to a point estimate, as proposed by \citet{Toccaceli2016}.
The suffixes -P and -VA indicate models calibrated with Platt scaling or Venn-ABERS predictors, respectively.
For instance, the calibrated MLPE model was labeled MLPE-P or MLPE-VA.

\subsection{Experiments}
\label{sec:experiments}
The first part of this study focuses on the data characteristics resulting from the temporal splitting strategy.
We studied the shift in label and descriptor space over time, mainly concentrating on the differences between TB and ADME-T assays.
The shift in label space was assessed by comparing the ratios of the preferred class in each time span.
Shifts in the descriptor space were quantified using the maximum mean discrepancy (MMD) \citep{Gretton2012}, which provides a kernel-based estimate for the distance between the ECFP spaces of two datasets.
The MMD between a training dataset $X := \{\bm{x_1}, ..., \bm{x_M}\}$ and a test dataset $Z := \{\bm{z_1}, ..., \bm{z_N}\}$ distributed according to $P(X)$ and $Q(Z)$ can be computed with

\begin{equation}
MMD(P,Q) = 
\frac{1}{M^2}\sum_{i=1}^{M}\sum_{j=1}^{M} k(\bm{x_{i}}, \bm{x_{j}}) + 
\frac{1}{N^2}\sum_{i=1}^{N}\sum_{j=1}^{N} k(\bm{z_{i}}, \bm{z_{j}}) -
\frac{1}{MN}\sum_{i=1}^{M}\sum_{j=1}^{N} k(\bm{x_{i}}, \bm{z_{j}}).    
\end{equation}

When using the Tanimoto coefficient \citep{Holliday2002} as kernel, the MMD lies between 0 and 1, with 0 indicating no differences and 1 indicating no shared features between the compounds in the datasets.

The second part of the study aims to determine how well different uncertainty quantification methods estimate the probability that compounds have a certain desirable feature, such as being active on a TB assay or inactive on an ADME-T toxicity assay.
To compare the models, the AUC under the receiver operating characteristic (ROC) curve \citep{Hanley1982} (AUC $[\uparrow]$), the binary cross-entropy (BCE $[\downarrow$]) and the adaptive calibration error \citep{Nixon2019} (ACE $[\downarrow$]) of the predictions were calculated.
The predictions were ordered and assigned to ten bins to obtain the calibration error.
Subsequently, the difference between the mean probabilities and the ratio of the instances belonging to the preferred class was computed for each bin.
The ACE was calculated by taking the mean of the differences in the bins.
Another commonly used calibration error is the expected calibration error, which is similar to the ACE but uses equally spaced instead of equally sized bins \citep{Nixon2019}.
However, this binning strategy can overestimate the calibration error when handling imbalanced datasets due to the high variance of the predictions in the sparsely populated bins \citep{Nixon2019}.
In this context, the ACE provides a more robust estimate of the calibration error and is, therefore, the preferred estimator for the probability calibration error in this study.
Note that the ACE is an improper score \citep{Gneiting2007, Brocker2009}, so a perfect calibration error of 0 does not automatically correspond to the best model.
Thus, the ACE was always evaluated in combination with the BCE for a more comprehensive analysis.
The performance of the approaches was assessed by generating ten model repetitions and obtaining the mean score of the respective evaluation metric.
A two-sided, independent t-test was used to assess whether the difference between the best and any other model was statistically significant.

\clearpage
\section{Results and Discussion}
\label{sec:results}
We divide the results of this study into two consecutive parts to investigate the efficacy of uncertainty quantification and probability calibration methods using real-world temporal data.
The first part addresses the properties of the data in the context of a distribution shift in the label and descriptor space.
In the second section, we compare the probability calibration of various uncertainty quantification methods and set the results in context with the underlying distribution shift resulting from the temporal split.

\subsection{Distribution Shifts over Time}
\label{sec:shifts}

\begin{figure}[t]
        \centering
        \includegraphics[width=\linewidth]{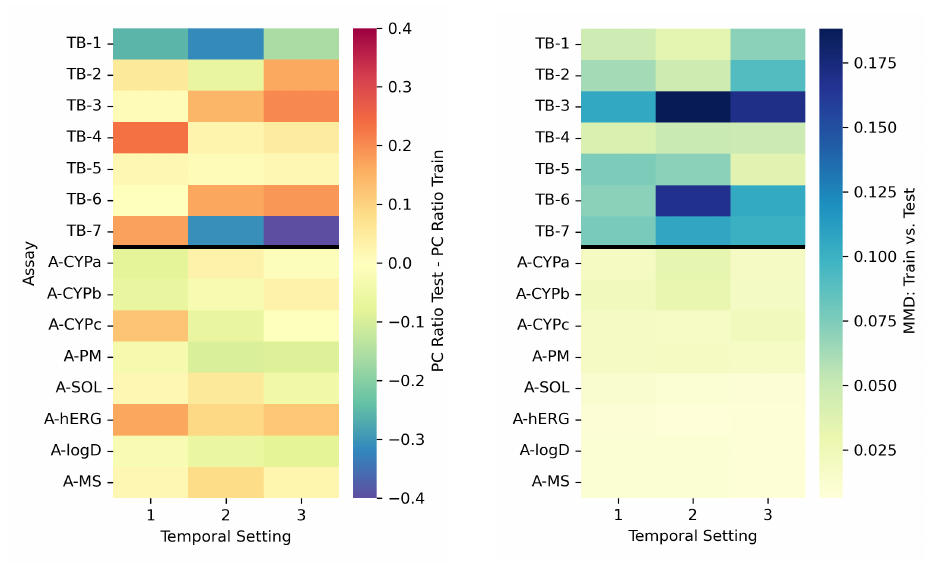}
        \caption{\textbf{Quantification of the distribution shifts between the training and test datasets over time.} The shift in label space and in the descriptor space is illustrated for each temporal setting, using the data of 1, 2, or 3 time spans for training. Results are shown for each assay. The left panel shows the shift in label space in terms of the difference in ratios of the preferred class between the training and test datasets. The right panel shows the MMD in the descriptor space between the training and test datasets for each temporal setting and assay, quantifying shifts in descriptor space.
        }
        \label{fig:shift}
    \end{figure}

\begin{figure}[h!]
        \centering
        \includegraphics[width=\linewidth]{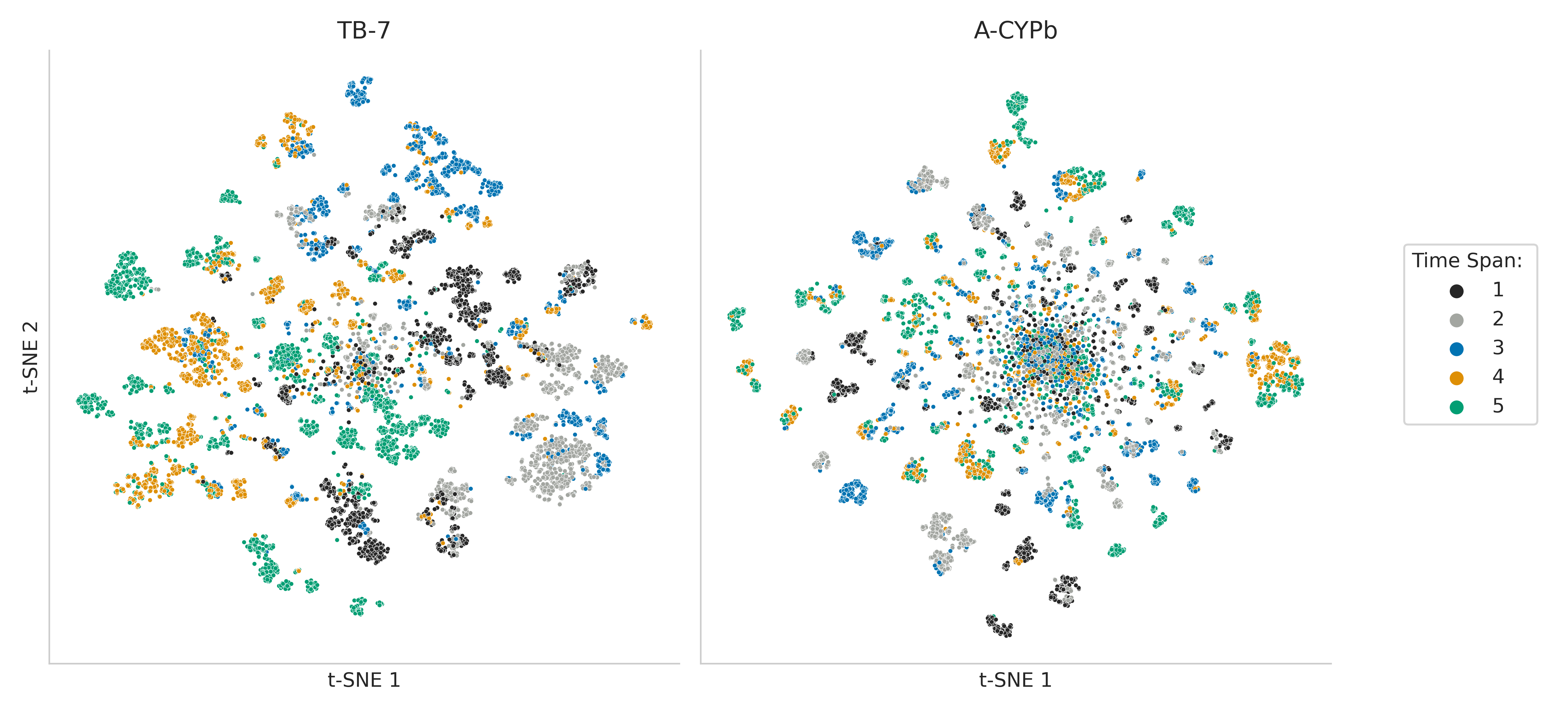}
        \caption{\textbf{T-SNE plots of the ECFP space.} T-SNE plots of the ECFP space are shown for one example of each assay category to illustrate how the explored chemical space changes over time. Compounds are colored according to the time span that they were assigned to. The t-SNE plot of the remaining TB and ADME-T assays are shown in Figures \ref{fig_supp:tsne_TB} and \ref{fig_supp:tsne_ADMET} in the appendix.
        }
        \label{fig:tsne}
    \end{figure}
\paragraph{Shift in Label Space.}
The ratios of the preferred class in different time spans of an assay were compared to evaluate shifts in the label space.
The left panel in Figures \ref{fig:shift} illustrates the distribution shift in label space by plotting the difference between the ratios of the preferred class in the training and the test set for each assay.
The ratio of the preferred class in each time span is listed for all assays in Table \ref{tab_supp:ratios} in the appendix.
We assessed all three temporal settings, using one, two, or three time spans as training data. 
The second consecutive span after the training data was considered the test set.
We refer to Figure \ref{fig:Split} for a more comprehensive explanation of the different training settings.
The left heatmap of Figure \ref{fig:shift} shows that TB assays evolve differently in label space over time than ADME-T assays.
Generally, the differences in preferred class ratios between the training and test sets are smaller in ADME-T assays, while the more extreme values in TB assays indicate larger shifts in label space.
Recall that the TB category includes project-based assays, which aim to find modulators for a specific target of interest.
Therefore, a plausible explanation for the larger shifts in label space could be that various chemical series are tested in search of promising compounds.
These series may differ in their abilities to modulate a target, leading to changing ratios of preferred compounds over time.
Some assays show an enrichment of the preferred class over time, as observed in TB-3 and TB-6.
However, this pattern was not observed in all assays, and the TB-1 and TB-7 assays even show opposite tendencies.
The more stable ratios in the ADME-T assays can be attributed to the nature of this assay category, as well. 
These assays are not specific to individual projects and are used to evaluate the pharmacokinetic and toxicity profiles of compounds.
Based on the results above, it is highly questionable whether the i.i.d. assumption for these assays remains valid over time, particularly in the TB category. 
This category includes some challenging assays, such as TB-3 and TB-7, which exhibit significant label shifts.

\paragraph{Shift in Descriptor Space.}

The shifts in ECFP space are visualized using two-dimensional t-SNE plots to reveal patterns and clusters in the dataset.
The t-SNE plots for TB-7 and A-CYPb are shown in Figure \ref{fig:tsne}, while the plots for the remaining assays can be found in Figures \ref{fig_supp:tsne_TB} and \ref{fig_supp:tsne_ADMET} in the appendix.
Figure \ref{fig:tsne} reveals a clear pattern in the TB-7 assay, which indicates a shift in the chemical space over the assay history.
Furthermore, chemically similar compounds tend to be assigned to the same time span, as indicated by the color purity in various clusters.
In contrast, the t-SNE plot of the ADME-T assay does not show a clear pattern in clusters and color gradients.
To quantify the shift in descriptor space, the MMD was calculated between the training and test set.
The MMD of the three temporal settings is shown in the right panel of Figure \ref{fig:shift}.
The observed tendencies are similar to the patterns reported for the label shifts.
In general, the TB assays exhibit larger shifts than the ADME-T assays, which is also supported by the patterns seen in the t-SNE plots.
These results can be again explained by the different characteristics of the two assay categories, resulting in distinct developments through the descriptor space over time.
To find promising compounds in the TB assays, various chemical series are usually screened, containing chemically similar compounds.
As a result, large shifts are observed when comparing the descriptor space of compounds assigned to different time spans.
In conclusion, the shifts in descriptor space are more pronounced in TB assays, while those in ADME-T assays are comparatively small. 
Similar to the shifts in label space, the i.i.d. assumption might not be appropriate, particularly in the TB assays.   

\subsection{Probability Calibration Study}
\label{sec:probcal}
We assessed the performance of various uncertainty estimation methods in three experiments, focusing on the probability calibration of the models.
Throughout this part of the study, the model performance is assessed separately for TB and ADME-T assays.
The first experiment compares the baseline approaches, and the uncertainty quantification approaches limited to the third temporal setting, in which three time spans were used as training datasets.
The second section investigates the change in model performances over time by comparing the three temporal settings. 
Moreover, the results will be linked to the assay-specific conclusions about the label and data shift drawn in Section \ref{sec:shifts}.
The third experiment concentrates on the potential of post hoc probability calibration methods in the context of distribution shifts between the calibration and test sets.
The results of the majority of models are presented. The numerical results of all methods are listed in Section \ref{sec_supp:calibrationstudy} in the appendix.

\paragraph{Comparison of Uncertainty Estimation Methods.}
\label{sec:UE}
We compared the predictive performance of RF, MLP, MLPE, MLPMC, and BNN in terms of AUC, BCE, and ACE.
For a straightforward comparison, we limit the reported results to the third temporal setting, in which three time spans were used for model training.
The data assigned to the last time span was used as a test set.
The AUC values of the models are listed in Table \ref{tab:auc}.
The AUC results for the TB assays show that the MLPs, as well as the non-bayesian uncertainty estimation methods, outperform the RFs and BNNs on most datasets.
In more detail, the MLPE model is always among the best approaches.
The MLP and the MLPMC models retrieve results that are not statistically different from the best in 6 and 5 out of 7 datasets.
The BNN is the best model for assay TB-4, while the RF approach is consistently outperformed.
In contrast, the results for the ADME-T assays show that either the MLPEs, or the BNNs, or both outperform the other approaches in 7 out of 8 assays.
The remaining assay is A-CYPa, for which the RF approach achieves the best result.

\begin{table*}
    \centering
    \caption{\textbf{Summary of AUC results for the third temporal setting.} AUC results for the baselines (RF and MLP) and train-time uncertainty quantification models (MLPE, MLPMC, and BNN) are reported. The models were trained with compounds from three time spans. The mean and standard deviation of 10 model repetitions are shown. The best-performing approach and those not significantly different from it, as determined by a two-sided t-test, are highlighted in bold.}
        \begin{tabular}{lrrrrr}
        \toprule
            & \multicolumn{1}{c}{RF} & \multicolumn{1}{c}{MLP} & \multicolumn{1}{c}{MLPE} & \multicolumn{1}{c}{MLPMC} & \multicolumn{1}{c}{BNN} \\
            \midrule
                \multicolumn{6}{l}{\textbf{Target-Based}}\\
                TB-1 & 0.437 $\pm$ 0.025 & \textbf{0.586 $\pm$ 0.065} & \textbf{0.614 $\pm$ 0.006} & \textbf{0.592 $\pm$ 0.061} & 0.502 $\pm$ 0.024 \\
                TB-2 & 0.778 $\pm$ 0.018 & \textbf{0.793 $\pm$ 0.01} & \textbf{0.795 $\pm$ 0.002} & \textbf{0.791 $\pm$ 0.01} & 0.289 $\pm$ 0.026 \\
                TB-3 & 0.708 $\pm$ 0.023 & \textbf{0.765 $\pm$ 0.009} & \textbf{0.768 $\pm$ 0.001} & 0.761 $\pm$ 0.009 & 0.73 $\pm$ 0.002 \\
                TB-4 & 0.909 $\pm$ 0.027 & 0.95 $\pm$ 0.007 & \textbf{0.956 $\pm$ 0.0} & 0.952 $\pm$ 0.003 & \textbf{0.957 $\pm$ 0.001} \\
                TB-5 & 0.676 $\pm$ 0.028 & \textbf{0.896 $\pm$ 0.008} & \textbf{0.9 $\pm$ 0.001} & \textbf{0.897 $\pm$ 0.008} & 0.78 $\pm$ 0.171 \\
                TB-6 & 0.641 $\pm$ 0.06 & \textbf{0.768 $\pm$ 0.007} & \textbf{0.771 $\pm$ 0.001} & \textbf{0.768 $\pm$ 0.007} & 0.701 $\pm$ 0.007 \\
                TB-7 & 0.533 $\pm$ 0.086 & \textbf{0.71 $\pm$ 0.026} & \textbf{0.718 $\pm$ 0.004} & \textbf{0.718 $\pm$ 0.03} & 0.479 $\pm$ 0.014 \\
            \midrule
                \multicolumn{6}{l}{\textbf{ADME-T}}\\
                A-CYPa & \textbf{0.675 $\pm$ 0.015} & 0.625 $\pm$ 0.013 & 0.627 $\pm$ 0.002 & 0.625 $\pm$ 0.013 & 0.622 $\pm$ 0.016 \\
                A-CYPb & 0.581 $\pm$ 0.012 & 0.644 $\pm$ 0.005 & 0.648 $\pm$ 0.001 & 0.643 $\pm$ 0.006 & \textbf{0.661 $\pm$ 0.001} \\
                A-CYPc & 0.631 $\pm$ 0.024 & 0.714 $\pm$ 0.003 & 0.714 $\pm$ 0.0 & 0.715 $\pm$ 0.003 & \textbf{0.734 $\pm$ 0.003} \\
                A-PM & 0.613 $\pm$ 0.031 & \textbf{0.769 $\pm$ 0.004} & \textbf{0.77 $\pm$ 0.001} & 0.765 $\pm$ 0.004 & \textbf{0.784 $\pm$ 0.023} \\
                A-SOL & 0.692 $\pm$ 0.008 & 0.78 $\pm$ 0.011 & \textbf{0.792 $\pm$ 0.002} & 0.779 $\pm$ 0.013 & 0.511 $\pm$ 0.011 \\
                A-hERG & 0.632 $\pm$ 0.009 & 0.729 $\pm$ 0.005 & \textbf{0.735 $\pm$ 0.001} & 0.729 $\pm$ 0.005 & 0.652 $\pm$ 0.003 \\
                A-logD & 0.641 $\pm$ 0.015 & 0.833 $\pm$ 0.003 & \textbf{0.839 $\pm$ 0.002} & 0.834 $\pm$ 0.003 & 0.828 $\pm$ 0.004 \\
                A-MS & 0.694 $\pm$ 0.006 & 0.71 $\pm$ 0.01 & 0.716 $\pm$ 0.003 & 0.713 $\pm$ 0.006 & \textbf{0.745 $\pm$ 0.007} \\
            \bottomrule
        \end{tabular}
        \label{tab:auc}
\end{table*}

\begin{figure}[t]
        \centering
        \includegraphics[width=\linewidth]{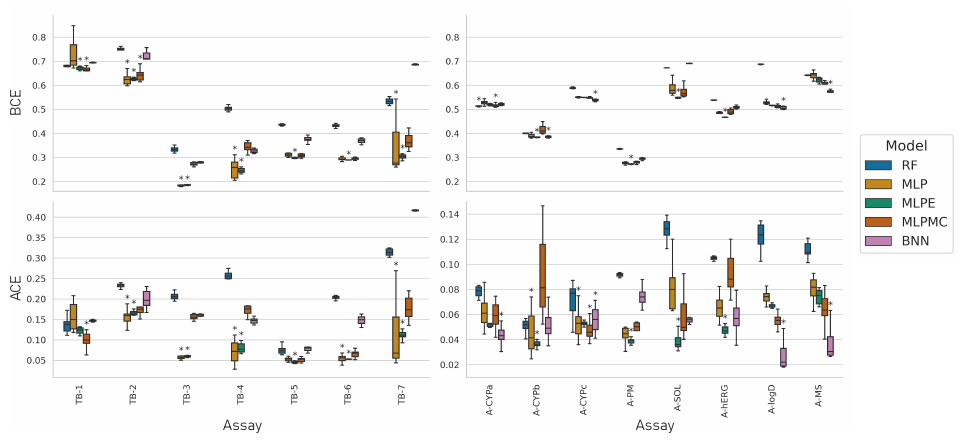}
        \caption{\textbf{Summary of BCE and ACE scores for the third temporal setting.} The first column shows the results for TB assays, while the second one reports the performance of models trained on ADME-T assays. BCE scores are plotted in the first, and ACE scores in the second row. Results for the baselines (RF and MLP) and train-time uncertainty quantification models (MLPE, MLPMC, and BNN) are reported. The models were trained with compounds from three time spans. The results of 10 model repetitions were aggregated. For each assay, the best-performing approach and those not significantly different from it, as determined by a two-sided t-test, are marked with an asterisk. }
        \label{fig:UE}
    \end{figure}

The model calibration is analyzed using the ACE and the BCE scores.
The results are illustrated in Figure \ref{fig:UE}.
The analysis of the BCE and ACE values reveals trends similar to those observed in the AUC scores.
The results of the models trained on TB assays show that the MLPE approach is among the best-performing approaches for all datasets, except for TB-1, for which MLPMC performs best in terms of ACE.
The baseline MLP matches the performance of MLPE in 4 out of 7 times in terms of BCE and in 5 out of 7 assays in terms of ACE.
Furthermore, the models perform overall worse on TB-1 and TB-2 in terms of BCE.
A reason for this result could be the small dataset size of these two assays, which might lead to overfitting and, therefore, poorer calibration of the models.
Furthermore, both assays exhibit comparatively large shifts in label and descriptor space, as shown in Figure \ref{fig:shift}, which introduces additional difficulties in generalizing well over time.
The results of the models trained on the ADME-T data demonstrate the superiority of the MLPE and BNN methods.
More specifically, in all assays, either MLPE or the BNN performs best with regard to BCE and ACE.
A-CYPa is the only exception for which RF and MLPMC perform best.
The MLPE and BNN models achieve the best results across both metrics in the same number of assays, namely in 4 out of 8 cases.

It is surprising that the more sophisticated uncertainty estimation methods improve the probability calibration of the baselines trained on AMDE-T assays but fail to do so for models trained on TB category datasets.
The MLPE and the BNN approaches account for epistemic uncertainty by including model variance in their predictions.
The deep ensemble approach retrieves model samples representing different local minima of the loss surface, while the BNNs model the neural network weights as probability distributions.
These characteristics should enable the models to detect out-of-distribution test instances that are dissimilar from the training instances.
As reported in Section \ref{sec:shifts}, the shifts innate to assays in the TB category are comparatively large, which leads to the assumption that these approaches enhance probability calibration for these assays. 
However, for most TB assays, the uncertainty estimation methods do not improve the quality of uncertainty estimates for the baseline MLPs.
\citet{Koh2021} showed that the performance of a classifier can degrade significantly when there is a distribution shift between the source and target domains. 
Moreover, \citet{Garg2020} reported specifically for the presence of label shift, that a classifier that is ideal for the source domain might no longer be ideal for the target domain.
In general, deep ensemble approaches have been shown to outperform other uncertainty estimation approaches, including approximate Bayesian neural networks, in terms of predictive accuracy and calibration without \citep{Gustafsson2020} and under distribution shift in the descriptor space \citep{Ovadia2019, Mehrtens2023}.
These conclusions are supported by the results in this study, that show that MLPE performs better than other uncertainty estimation under distribution shift.
However, despite being the best uncertainty estimation approach, the MLPE models rarely outperform the baseline MLP in the presence of distribution shift.
An additional reason for the failure of the uncertainty estimation methods to generate better uncertainty estimates that the baseline could be their inability to handle the shifts in label space well.
This conclusion might be transferrable to model calibration, thus explaining the difference between architectures trained on TB and ADME-T assays to produce better-calibrated probabilities.
A large study that compared uncertainty estimation approaches used for regression models trained on the same dataset also reported that the deep ensemble and Bayesian neural network approaches outperform other common uncertainty estimation approaches for regression \citep{Svensson2024arxiv}.
In contrast to the classification setting, the uncertainty estimates of baselines trained with TB assays could also be improved in the regression study.
This observation could indicate that uncertainty estimation models for regression tasks are less sensitive to shifts in the label space than approaches for classification. 
A reason for this observation could be that the model can access the actual values of the measurements in the regression setting, which might attenuate the shift in the target space.
For example, strongly inactive and weakly inactive compounds exhibit different target values for regression models, while this information is lost in classification tasks due to the application of binary classification thresholds.

\paragraph{Uncertainty Estimation over Time.}
\label{sec:UEovertime}

\begin{figure}[t]
        \centering
        \includegraphics[width=\linewidth]{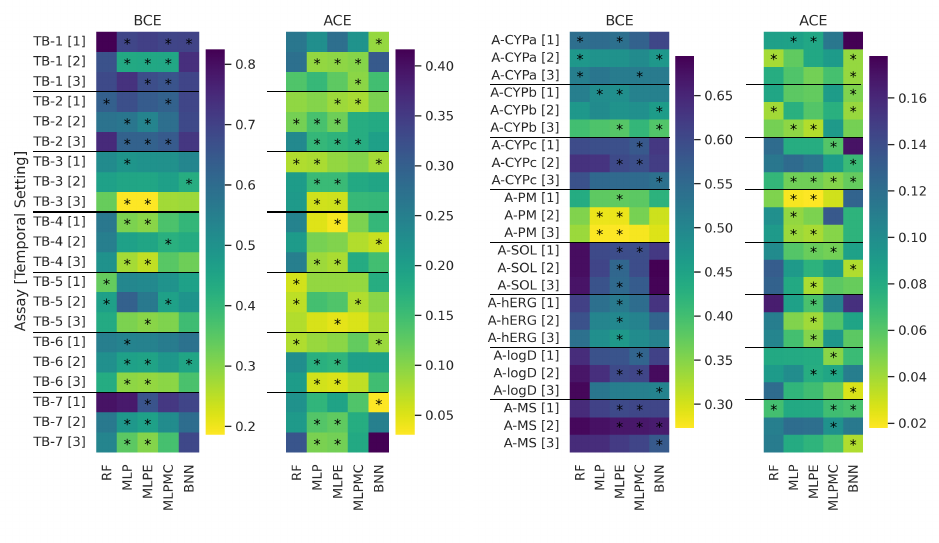}
        \caption{\textbf{Summary of BCE and ACE scores across all temporal settings.} The first two columns show the ACE and BCE scores for TB assays, while the last two report the performance of models trained on ADME-T assays. The temporal setting is indicated in brackets after the assay abbreviation. Results for the baselines (RF and MLP) and train-time uncertainty quantification models (MLPE, MLPMC, and BNN) are reported. The models were trained with compounds from three time spans. Averages over 10 model repetitions are shown. For each assay, the best-performing approach and those not significantly different from it, as determined by a two-sided t-test, are marked with an asterisk. }
        \label{fig:UEovertime}
    \end{figure}
    
We assessed the quality of the uncertainty estimates over time by comparing the model performance in all three temporal settings.
The results are displayed in Figure \ref{fig:UEovertime}.
The plots show that the model performance in one time span rarely allows conclusions about the performance of the same approach at another point in time.
In 3 out of 7 TB assays, a single model is always among the models with the best BCE score in all three temporal settings.
These approaches include MLPMC for TB-1, MLP for TB-6, and MLPE for TB-7. 
Regarding the ACE scores, the MLPE is consistently among the best methods for TB-2, and the MLP is among the best for TB-3.
The BCE results of the ADME-T assays show consistent results over time for 4 assays, including the RFs for A-CYPa and the MLPE approach for A-PM, A-SOL, and A-hERG. 
The ACE results of the ADME-T assays reveal a consistent model performance in 2 out of 8 assays, namely the MLPs for A-PM and the MLPE for A-hERG.
In general, model performance in terms of ACE is slightly less stable 
over time than in terms of BCE.
Interestingly, \citet{Svensson2024arxiv} applied regression modeling techniques to the same data and reported more consistent model performance for the ADME-T assays.

Plotting the model performance on all temporal settings confirms the conclusions drawn in Section \ref{sec:UE}.
The MLP and MLPE methods obtain the best results in terms of both BCE and ACE for the TB assays.
Both are among the best-performing approaches in at least one temporal setting in all assays, except TB-4, where MLP is not among the best models in any setting.
Interestingly, regarding the BCE results, the MLP is as often among the best models as the MLPEs, namely for 13 out of the 21 settings.
The ACE scores reveal a similar result.
MLP is among the best methods for 11 and MLPE for 13 out of 21 settings.
All other approaches trained on the TB assays are much less often among the best-performing models.
The results for the ADME-T assays also support the results from the previous experiment.
Concerning the ACE scores, the MLPE and BNNs outperform all other models, with MLPE being among the best methods for 13 and BNN for 11 out of 24 settings.
The MLPs and the MLPMCs are among the best-calibrated methods in 5 and 7 settings, respectively.
The MLPEs outperform the other approaches in terms of BCE, obtaining significant results in 13 settings.
The MLPMCs are among the best methods in 8, and the BNNs in 7 out of 24 settings.

In conclusion, the MLPE approach generates the best-performing models for most ADME-T assays.
Considering the high computational resources required to train these MLPEs, another good choice are BNNs. 
This approach is much more time—and resource-effective and generates well-calibrated models for many ADME-T datasets.
Note that a fixed variance was chosen for the Gaussian prior and that tuning this hyperparameter could even improve the performance of the BNNs.
Given that deep ensemble models always lead to more underconfident predictions \citep{Rahaman2021}, the good performance of these methods indicates overfitting of the baselines.
The MLPEs and BNNs counteract this behavior for the ADME-T assays but not for datasets from the TB category.
The ACE values for TB assays are higher than for the ADME-T assays, as illustrated in Figure \ref{fig:UE}, indicating worse model calibration.
This result indicates room for improvement in probability calibration and eliminates good baseline model calibration as a reason for the inability of MLPE and BNN to improve the ACE of the MLP.
An alternative reason for the failure of MLPEs and BNNs could be the shift in label space which is considerably larger for TB assays.
Based on these results, it is questionable whether the costly generation of MLPEs for improved uncertainty estimation is justified for datasets with large distribution shifts, such as the TB data.
As discussed in Section \ref{sec:UE}, this shift might be difficult to handle for the uncertainty quantification methods, resulting in the incapability of uncertainty quantification methods to improve the probability calibration.  

\paragraph{Post hoc Probability Calibration.}

\begin{figure}[t]
        \centering
        \includegraphics[width=\linewidth]{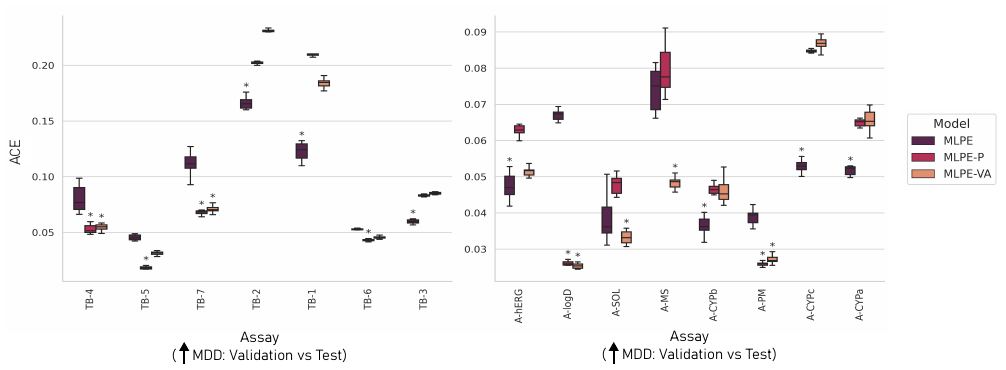}
        \caption{\textbf{Summary of ACE scores of post-hoc probability calibration approaches using the third temporal setting.} The left panel shows the ACE scores of models trained on TB assays, while the right panel reports the ACE performance of ADME-T models. The assays in each panel are ordered according to increasing distribution shifts in descriptor space between the calibration and the training set. The distribution shift is determined by the maximum mean discrepancy (MMD) between the two datasets using the Tanimoto coefficient on the ECFP space determines the distribution shift. Results for the deep ensembles (MLPE), the Platt scaled deep-ensembles (MLPE-P), and the ensembles calibrated with a Venn-ABERS predictor (MLPE-VA) are reported. The models were trained with compounds from three time spans. The results of 10 model repetitions were aggregated. For each assay, the best-performing approach and those not significantly different from it, as determined by a two-sided t-test, are marked with an asterisk. }
        \label{fig:probcal}
    \end{figure}

The effects of two post hoc calibration methods, Platt scaling and Venn-ABERS predictors, were assessed.
Furthermore, the distribution shift between calibration and test set was taken into account by obtaining the MMD between the two datasets.
The MMD between the calibration and test dataset is reported in Table \ref{tab_supp:MMD_val} in the appendix.
For the sake of clarity, we only include the results of the third temporal setting for the MLPE approach, which was reported to be one of the best-performing models in previous sections of this study.
Only the ACE is reported since the post hoc calibration step includes the application of monotonous increasing functions, which cannot correct non-monotonous distortions and, therefore, does not change the ranking of the predictions. 
Hence, these approaches only affect the probability calibration while the AUC scores of the models remain constant. 
The results of all calibrated models across all temporal settings and assays can be found in Section \ref{sec_supp:calibrationstudy} of the appendix.
Figure \ref{fig:probcal} shows the performance of the Platt-scaled MLPEs (MLPE-P) and the MLPEs calibrated with Venn-ABERS predictors (MLP-VA).
The results are shown separately for TB and ADME-T assays, and the assays are ordered according to increasing MMD values.
In general, post hoc scaling leads to better results in 4 out of 7 TB assays and in 4 out of 8 ADME-T assays.
In 4 out of these 8 cases, both MLPE-P and MLPE-VA are the best models, while in 4 cases, either MLPE-P or MLPE-VA perform best.
Both panels in Figure \ref{fig:probcal} show, that with increasing MMD between calibration and test set, the calibrating abilities of the post hoc calibration methods decrease.
A stronger trend can be detected for the TB assays, for which the shifts are larger than for the ADME-T assays.
Nevertheless, the pattern is also visible in the right panel plotting the results for ADME-T, which show no improvements after post hoc calibration for assays with large MMD, like A-CYPc and A-CYPa.
The reported trends can also be seen for other approaches, albeit less clearly for some models. 
An intuitive explanation for this pattern is that post hoc probability approaches perform better if the training set is similar to the test set, and worse if the training and test set are different.
\citet{Ovadia2019} showed that post hoc calibration approaches improve model calibration in the i.i.d. setting.
However, their calibrating abilities degrade as the data shift increases so that they are ultimately outperformed by train-time uncertainty quantification methods in the presence of large shifts.
These findings are supported by the post hoc calibration results in this study, which shows for small shifts good calibrating properties of the post hoc calibration methods, while for larger shifts, they are outperformed by the train-time versions of the methods.

\section{Conclusions}
\label{sec:conclusions}
Uncertainty estimation emerges as a critical tool in the cost- and resource-intensive drug discovery process, facilitating the evaluation of experimental risks and costs. 
In this framework, the quality of the uncertainty estimates is crucial to ensure the reliability of the models.
This study evaluates uncertainty estimation approaches for classification tasks in a practical, real-world context.
A temporal splitting strategy was applied to internal data from a pharmaceutical company to simulate the evolution of drug-target interaction data through time.  
Our findings offer valuable insights into uncertainty estimation and highlight the challenges posed by real-world applications.

The analysis of the pharmaceutical data showed that the distribution shifts in label and descriptor space over time strongly depended on the nature of the individual assays.
The project-specific TB assays exhibited more pronounced distribution shifts, while the more abundantly used toxicity screens and assays assessing the pharmacodynamic properties (ADME-T assays) showed moderate and more stable shifts in descriptor space and little shift in label space over time.
These results suggest that the i.i.d. assumption might not be accurate, especially for target-specific assays with larger shifts in label- and descriptor space.

A comparison of common uncertainty quantification methods revealed that the deep ensembles and the Bayesian neural network achieved the best-calibrated results for ADME-T assays that show small distribution shifts in the data.
Recently, these two approaches have also been shown to outperform other common uncertainty quantification methods in regression tasks \citep{Svensson2024arxiv}.
Given that training deep ensembles demand a lot of computational resources, Bayesian neural networks might be the best choice when striving for a fast method that produces well-calibrated estimates.
For TB assays exhibiting more pronounced distribution shifts, the deep ensemble method performed best.
However, the baseline MLP matched the performance of the uncertainty quantification methods for many datasets.
In general, the uncertainty quantification methods consider epistemic uncertainty, resulting in less confident predictions for test instances that are different from the training data.
This leads to the assumption that the inability of the uncertainty estimation approaches to produce well-calibrated uncertainties for TB assays might result from the shifts in label space rather than shifts in descriptor space.
However, due to the high computational effort required to train deep ensembles, choosing a simple MLP for assays with large distribution shifts might be the best and most efficient solution.

The analysis of the performance of uncertainty estimation approaches over time showed that it is difficult to draw conclusions from results from one point of time in the assay history to another.
In general, model performance was unstable over time.
Only for a few assays could one method be identified that was among the best-performing approaches at all considered time points in the assay history.
In conclusion, a reevaluation of classification model performance is required for all assays as soon as more recent data is added.
Interestingly, the performance of regression models is more stable over time for ADME-T assays, as shown recently by \citet{Svensson2024arxiv}.

Lastly, two post hoc calibration approaches, including Platt scaling and Venn-ABERS predictors, were tested on their ability to improve the probability calibration of deep ensembles.
Overall, the calibration methods were able to achieve better-calibrated results for some assays, while for others, the calibration stayed the same or even deteriorated after the post hoc calibration step.
The calibrating capabilities were dependent on the distribution shift between the calibration and test set, with declining performance when the shifts in the descriptor space increase.

The uncertainty estimation approaches discussed in this study have been previously demonstrated to improve model calibration on toy data or datasets that do not account for the distribution shift caused by the evolution of the data over time.
However, previous studies show that improved performance on i.i.d. data often fails to translate into better outcomes under distribution shift \citep{Ovadia2019, Koh2021}, which is also supported by the findings in this study.
This study highlights the challenges introduced by real-world data, emphasizing the complexity of identifying effective strategies for uncertainty estimation in QSAR models.

\section*{Acknowledgements}

The authors sincerely thank their colleagues for the useful discussions. Special appreciation goes to András Formanek and Antoine Passemiers at KU Leuven for their valuable insights. This study was partially funded by the European Union’s Horizon 2020 research and innovation programme under the Marie Skłodowska-Curie Actions grant agreement “Advanced machine learning for Innovative Drug Discovery (AIDD)” No. 956832. HRF, and AA are affiliated to Leuven.AI and received funding from the Flemish Government (AI Research Program).

\clearpage
\bibliographystyle{unsrtnat}
\bibliography{bibliography}

\newpage
\appendix
\section{Model Selection}
\label{sec_supp:model_selection}
\begin{table}[h!]
    \caption{\textbf{Summary of the hyperparameter optimization strategy.} The hyperparameters of the baseline models were tuned for each temporal setting of each dataset individually using the BCE loss in a validation dataset. The range of the hyperparameters considered during the hyperparameter tuning is reported in the left column. The right column lists the fixed hyperparameters for each baseline model.}
    \label{tab_supp:hyperparameters}
    \vskip 0.15in
    \centering
        \begin{adjustbox}{max width=\textwidth}
            \begin{tabular}{llr|lr}
            \toprule
                Baseline & Tuned Hyperparameter & Explored Values & Fixed Hyperparameter & Values \\
            \midrule
                \multirow{2}{7em}{Random Forest (RF)} & Number of Estimators & 50 - 1500 & \multirow{2}{*}{ECFP Size} & \multirow{2}{*}{4096}\\
                & Maximum Depth & 5 - 10000&&\\
            \midrule
                \multirow{6}{7em}{Mulitlayer Perceptron (MLP)} & Weight Decay & 0 - 0.0005 & &\\
                & Dropout & 0 - 0.75 &&\\
                & Hidden Dimension & 64 - 512 & ECFP Size & 4096 \\
                & Number of Layers & 2 - 5 & Learning Rate & 0.0001\\
                & Decreasing Dimension & True$/$False &&\\
                & Scheduler Factor & 0.1$/$0.5 &&\\
            \bottomrule
    \end{tabular}
    \end{adjustbox}
    \end{table}

\section{Distribution Shift in Label Space}
\label{sec_supp:distrshiftlabel}
    
\begin{table*}[h!]
    \caption{\textbf{Overview of ratio of the preferred class.} The ratios of the compounds assigned to the preferred class are reported for each time span and each assay.}
    \label{tab_supp:ratios}
    \centering
    \begin{tabular}{llrrrrr}
        \toprule
            Assay & Span 1 & Span 2 & Span 3 & Span 4 & Span 5 \\
        \midrule
            TB-1 & 0.678218 & 0.531628 & 0.422665 & 0.236022 & 0.355208 \\
            TB-2 & 0.336185 & 0.401515 & 0.389691 & 0.304681 & 0.537486 \\
            TB-3 & 0.780019 & 0.687738 & 0.789397 & 0.878903 & 0.955903 \\
            TB-4 & 0.547912 & 0.869677 & 0.779412 & 0.715577 & 0.760392 \\
            TB-5 & 0.102444 & 0.245645 & 0.119409 & 0.176832 & 0.168000 \\
            TB-6 & 0.750672 & 0.657795 & 0.747870 & 0.870065 & 0.905208 \\
            TB-7 & 0.299052 & 0.717688 & 0.472853 & 0.189217 & 0.079219 \\
        \midrule
            A-CYPa & 0.853501 & 0.675146 & 0.775178 & 0.817248 & 0.773893 \\
            A-CYPb & 0.856396 & 0.844581 & 0.793540 & 0.823023 & 0.864290 \\
            A-CYPc & 0.606965 & 0.761506 & 0.728951 & 0.635174 & 0.707853 \\
            A-PM & 0.186431 & 0.195426 & 0.155797 & 0.096440 & 0.090965 \\
            A-SOL & 0.470625 & 0.472610 & 0.488615 & 0.525761 & 0.434304 \\
            A-hERG & 0.559480 & 0.694663 & 0.726091 & 0.719838 & 0.783182 \\
            A-logD & 0.633528 & 0.600615 & 0.608635 & 0.554792 & 0.533047 \\
            A-MS & 0.340375 & 0.333128 & 0.356558 & 0.421075 & 0.365665 \\
        \bottomrule
    \end{tabular}
\end{table*}

\newpage
\section{Distribution Shift in Descriptor Space}
\label{sec_supp:distrshiftdescriptor}

\begin{figure*}[h!]
    \vskip 0.2in
    \begin{center}
    \centerline{\includegraphics[width=\textwidth]{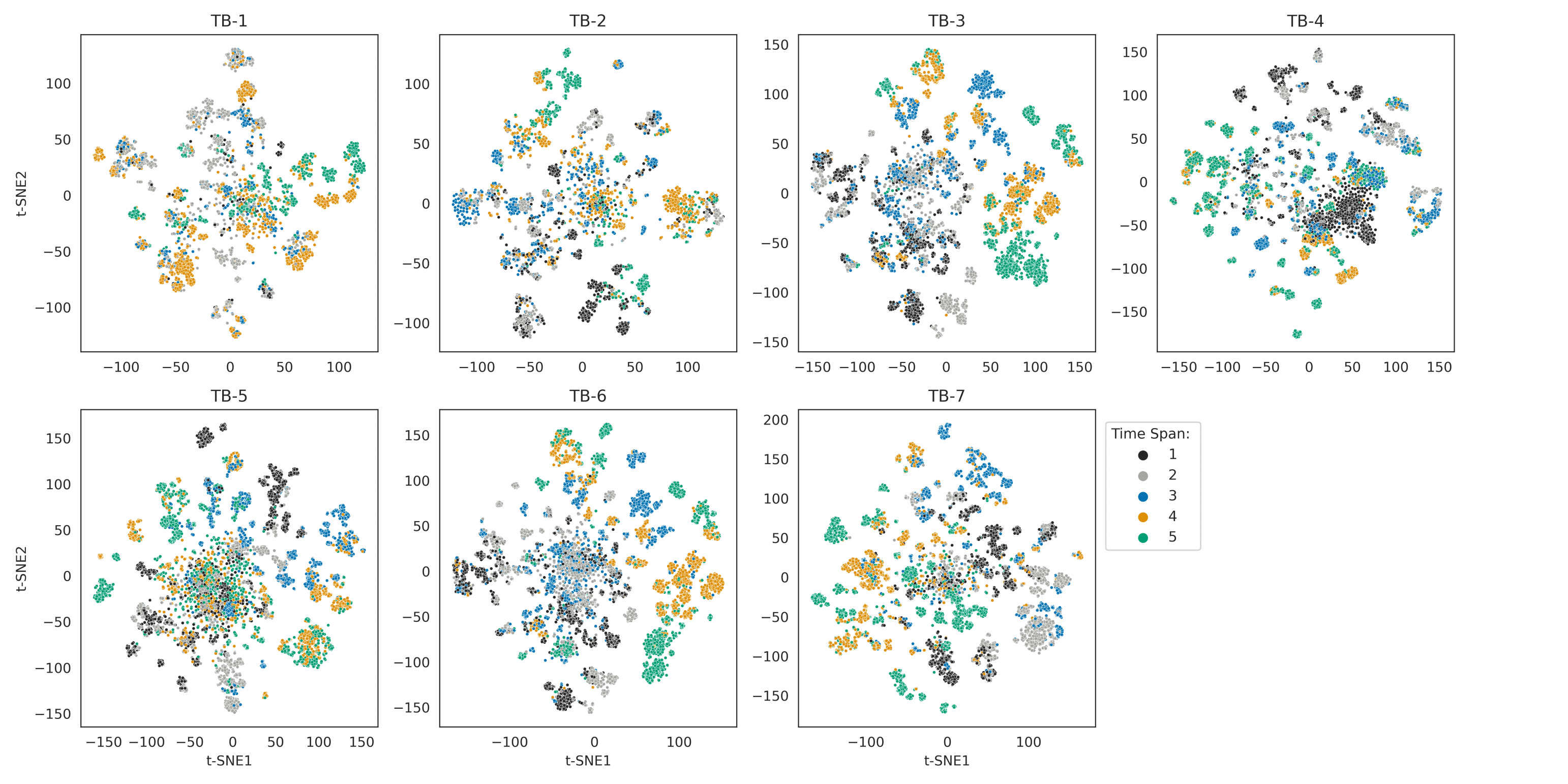}}
     \caption{\textbf{T-SNE plots of the ECFP space for all TB assays.} T-SNE plots of the ECFP space are shown for all TB assays. Compounds are colored according to the time span that they were assigned to.
        }
    \label{fig_supp:tsne_TB}
    \end{center}
    \vskip -0.2in
\end{figure*}

\begin{figure*}[h!]
    \vskip 0.2in
    \begin{center}
    \centerline{\includegraphics[width=\textwidth]{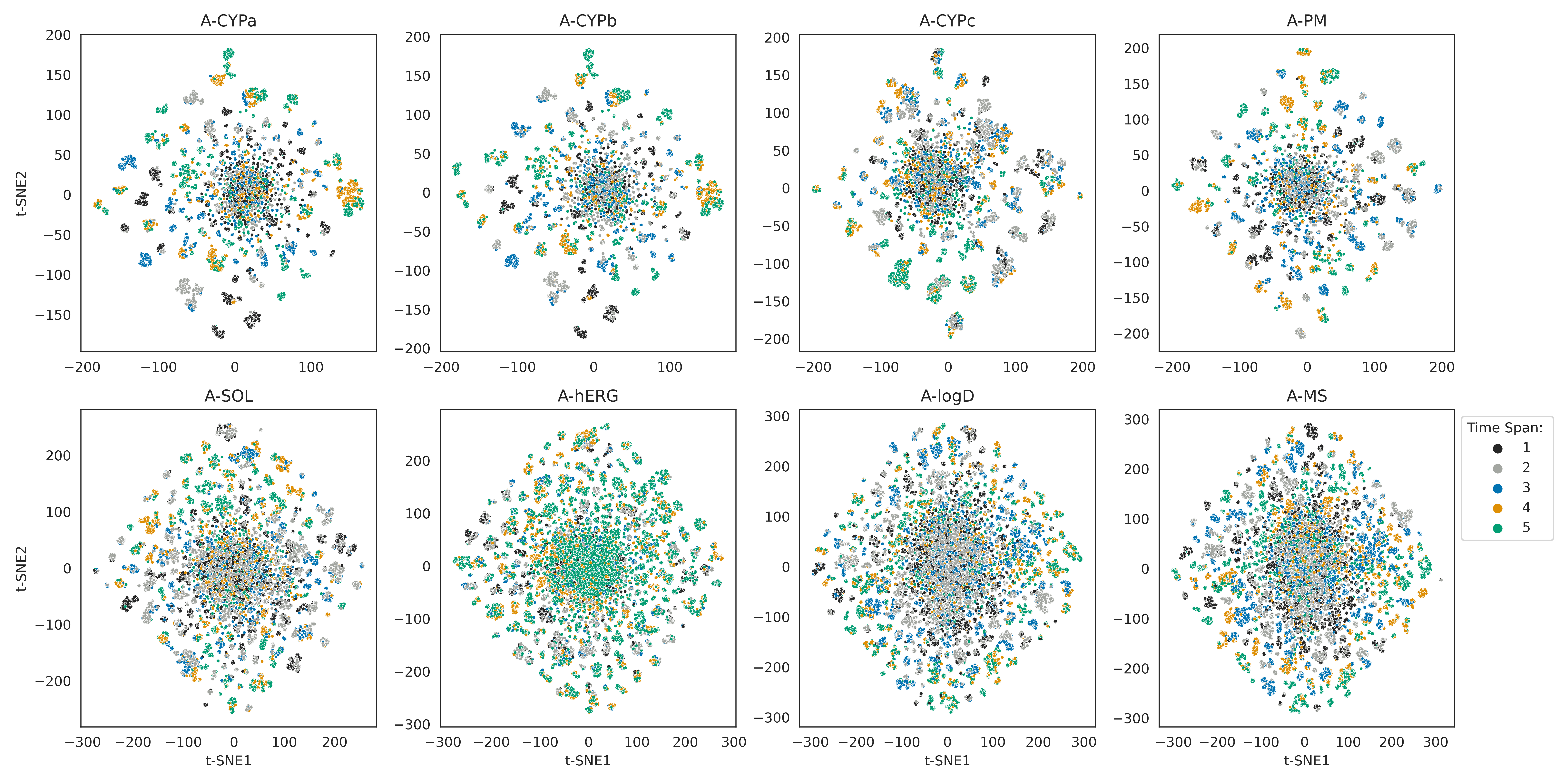}}
     \caption{\textbf{T-SNE plots of the ECFP space for all ADME-T assays.} T-SNE plots of the ECFP space are shown for all ADME-T assays. Compounds are colored according to the time span that they were assigned to.
        }
    \label{fig_supp:tsne_ADMET}
    \end{center}
    \vskip -0.2in
\end{figure*}

\newpage
\begin{table}[h!]
     \caption{\textbf{Quantification of the label distribution shifts between the calibration and test set.} The shift in label space is reported for each temporal setting, using the data of 1, 2, or 3 time spans for training. Results are shown for each assay. The shift in label space is shown in terms of the difference in ratios of the preferred class between the training and test datasets.
        }
    \label{tab_supp:MMD_val}
    \centering
         \begin{tabular}{lrrr}
            \toprule
                & \multicolumn{3}{c}{Number of Training Spans}\\
            \cmidrule{2-4}
                Assay & 1 & 2 & 3 \\
            \midrule
            \textbf{TB}\\
                TB-1 & 0.022208 & 0.016968 & 0.057858 \\
                TB-2 & 0.018045 & 0.061332 & 0.050021 \\
                TB-3 & 0.063842 & 0.098653 & 0.073235 \\
                TB-4 & 0.033499 & 0.043690 & 0.020273 \\
                TB-5 & 0.088506 & 0.054487 & 0.035291 \\
                TB-6 & 0.044372 & 0.107566 & 0.062775 \\
                TB-7 & 0.130317 & 0.095124 & 0.048923 \\
            \midrule
            \textbf{ADME-T}\\       
                A-CYPa & 0.024293 & 0.028514 & 0.016945 \\
                A-CYPb & 0.018977 & 0.026954 & 0.013106 \\
                A-CYPc & 0.010841 & 0.010273 & 0.016587 \\
                A-PM & 0.015786 & 0.016645 & 0.015720 \\
                A-SOL & 0.006665 & 0.008250 & 0.007077 \\
                A-hERG & 0.004232 & 0.003890 & 0.006738 \\
                A-logD & 0.008009 & 0.005819 & 0.007059 \\
                A-MS & 0.008084 & 0.006091 & 0.007487 \\        
             \bottomrule
    \end{tabular}
\end{table}

\section{Results of the Calibration Study}
\label{sec_supp:calibrationstudy}

\fontsize{9pt}{9pt}\selectfont
\begin{adjustbox}{angle=0, caption = {\textbf{Summary of the BCE ($\downarrow$) scores for the baseline models}. BCE results for all three temporal settings of all TB and ADME-T assays are reported. The temporal setting is indicated in brackets after the assay abbreviation.
Results for the baselines (Random Forests (RF) and a multilayer perceptron (MLP)),  Platt scaled models (RF-P and MLP-P) and models calibrated with Venn-ABERS (VA) predictors (RF-VA and MLP-VA) are reported.
The models were trained with compounds from three time spans. Averages over 10 model repetitions are shown. 
} , float=table}
    \label{tab_supp:res_bl_bce}
    \centering
    \begin{tabular}{lrrrrrr}
        \toprule
            Assay & RF & RF-P & RF-VA & MLP & MLP-P & MLP-VA \\
        \midrule
            TB-1[1] & 0.82 $\pm$ 0.012 & 0.7 $\pm$ 0.006 & 0.7 $\pm$ 0.006 & 0.69 $\pm$ 0.017 & 0.64 $\pm$ 0.055 & 0.6 $\pm$ 0.064 \\
            TB-1[2] & 0.66 $\pm$ 0.005 & 0.51 $\pm$ 0.006 & 0.51 $\pm$ 0.012 & 0.44 $\pm$ 0.012 & 0.43 $\pm$ 0.004 & 0.44 $\pm$ 0.005 \\
            TB-1[3] & 0.68 $\pm$ 0.003 & 0.81 $\pm$ 0.036 & 0.77 $\pm$ 0.011 & 0.73 $\pm$ 0.066 & 0.77 $\pm$ 0.03 & 0.77 $\pm$ 0.036 \\
            TB-2[1] & 0.63 $\pm$ 0.007 & 0.61 $\pm$ 0.009 & 0.63 $\pm$ 0.015 & 0.67 $\pm$ 0.038 & 0.67 $\pm$ 0.027 & 0.64 $\pm$ 0.006 \\
            TB-2[2] & 0.58 $\pm$ 0.003 & 0.56 $\pm$ 0.005 & 0.56 $\pm$ 0.006 & 0.57 $\pm$ 0.062 & 0.54 $\pm$ 0.011 & 0.54 $\pm$ 0.009 \\
            TB-2[3] & 0.75 $\pm$ 0.009 & 0.8 $\pm$ 0.025 & 0.78 $\pm$ 0.024 & 0.62 $\pm$ 0.024 & 0.66 $\pm$ 0.011 & 0.7 $\pm$ 0.013 \\
            TB-3[1] & 0.52 $\pm$ 0.003 & 0.64 $\pm$ 0.025 & 0.68 $\pm$ 0.016 & 0.49 $\pm$ 0.009 & 0.56 $\pm$ 0.005 & 0.56 $\pm$ 0.008 \\
            TB-3[2] & 0.46 $\pm$ 0.011 & 0.38 $\pm$ 0.011 & 0.37 $\pm$ 0.009 & 0.45 $\pm$ 0.016 & 0.44 $\pm$ 0.006 & 0.52 $\pm$ 0.093 \\
            TB-3[3] & 0.33 $\pm$ 0.01 & 0.2 $\pm$ 0.009 & 0.2 $\pm$ 0.013 & 0.19 $\pm$ 0.006 & 0.21 $\pm$ 0.006 & 0.21 $\pm$ 0.004 \\
            TB-4[1] & 0.57 $\pm$ 0.013 & 0.39 $\pm$ 0.043 & 0.4 $\pm$ 0.067 & 0.31 $\pm$ 0.018 & 0.31 $\pm$ 0.016 & 0.31 $\pm$ 0.007 \\
            TB-4[2] & 0.55 $\pm$ 0.005 & 0.45 $\pm$ 0.02 & 0.47 $\pm$ 0.011 & 0.46 $\pm$ 0.023 & 0.46 $\pm$ 0.029 & 0.47 $\pm$ 0.021 \\
            TB-4[3] & 0.5 $\pm$ 0.012 & 0.33 $\pm$ 0.031 & 0.32 $\pm$ 0.037 & 0.27 $\pm$ 0.082 & 0.24 $\pm$ 0.028 & 0.24 $\pm$ 0.038 \\
            TB-5[1] & 0.34 $\pm$ 0.003 & 0.41 $\pm$ 0.005 & 0.41 $\pm$ 0.021 & 0.53 $\pm$ 0.041 & 0.39 $\pm$ 0.005 & 0.37 $\pm$ 0.003 \\
            TB-5[2] & 0.46 $\pm$ 0.007 & 0.48 $\pm$ 0.018 & 0.49 $\pm$ 0.018 & 0.63 $\pm$ 0.076 & 0.53 $\pm$ 0.018 & 0.52 $\pm$ 0.024 \\
            TB-5[3] & 0.44 $\pm$ 0.004 & 0.42 $\pm$ 0.008 & 0.41 $\pm$ 0.011 & 0.31 $\pm$ 0.007 & 0.28 $\pm$ 0.008 & 0.29 $\pm$ 0.006 \\
            TB-6[1] & 0.55 $\pm$ 0.002 & 0.64 $\pm$ 0.026 & 0.67 $\pm$ 0.029 & 0.54 $\pm$ 0.004 & 0.65 $\pm$ 0.006 & 0.66 $\pm$ 0.007 \\
            TB-6[2] & 0.51 $\pm$ 0.011 & 0.43 $\pm$ 0.011 & 0.43 $\pm$ 0.02 & 0.46 $\pm$ 0.036 & 0.44 $\pm$ 0.017 & 0.5 $\pm$ 0.062 \\
            TB-6[3] & 0.43 $\pm$ 0.007 & 0.3 $\pm$ 0.012 & 0.3 $\pm$ 0.012 & 0.29 $\pm$ 0.007 & 0.28 $\pm$ 0.004 & 0.29 $\pm$ 0.003 \\
            TB-7[1] & 0.79 $\pm$ 0.009 & 0.73 $\pm$ 0.024 & 0.69 $\pm$ 0.032 & 0.78 $\pm$ 0.107 & 0.72 $\pm$ 0.133 & 0.73 $\pm$ 0.125 \\
            TB-7[2] & 0.57 $\pm$ 0.007 & 0.54 $\pm$ 0.019 & 0.46 $\pm$ 0.036 & 0.48 $\pm$ 0.041 & 0.47 $\pm$ 0.028 & 0.44 $\pm$ 0.016 \\
            TB-7[3] & 0.53 $\pm$ 0.013 & 0.32 $\pm$ 0.013 & 0.34 $\pm$ 0.031 & 0.35 $\pm$ 0.113 & 0.28 $\pm$ 0.009 & 0.28 $\pm$ 0.011 \\
            \midrule
            A-CYPa[1] & 0.54 $\pm$ 0.003 & 0.53 $\pm$ 0.005 & 0.52 $\pm$ 0.01 & 0.55 $\pm$ 0.01 & 0.55 $\pm$ 0.007 & 0.57 $\pm$ 0.023 \\
            A-CYPa[2] & 0.47 $\pm$ 0.002 & 0.48 $\pm$ 0.003 & 0.5 $\pm$ 0.014 & 0.48 $\pm$ 0.004 & 0.48 $\pm$ 0.002 & 0.5 $\pm$ 0.005 \\
            A-CYPa[3] & 0.51 $\pm$ 0.003 & 0.51 $\pm$ 0.005 & 0.51 $\pm$ 0.005 & 0.53 $\pm$ 0.009 & 0.53 $\pm$ 0.004 & 0.52 $\pm$ 0.005 \\
            A-CYPb[1] & 0.51 $\pm$ 0.003 & 0.5 $\pm$ 0.007 & 0.49 $\pm$ 0.007 & 0.49 $\pm$ 0.008 & 0.49 $\pm$ 0.003 & 0.49 $\pm$ 0.002 \\
            A-CYPb[2] & 0.47 $\pm$ 0.001 & 0.53 $\pm$ 0.01 & 0.56 $\pm$ 0.008 & 0.49 $\pm$ 0.015 & 0.51 $\pm$ 0.012 & 0.52 $\pm$ 0.02 \\
            A-CYPb[3] & 0.4 $\pm$ 0.001 & 0.4 $\pm$ 0.001 & 0.4 $\pm$ 0.002 & 0.39 $\pm$ 0.008 & 0.39 $\pm$ 0.001 & 0.39 $\pm$ 0.002 \\
            A-CYPc[1] & 0.6 $\pm$ 0.002 & 0.56 $\pm$ 0.005 & 0.57 $\pm$ 0.014 & 0.59 $\pm$ 0.009 & 0.58 $\pm$ 0.003 & 0.58 $\pm$ 0.002 \\
            A-CYPc[2] & 0.63 $\pm$ 0.005 & 0.62 $\pm$ 0.012 & 0.6 $\pm$ 0.017 & 0.63 $\pm$ 0.027 & 0.6 $\pm$ 0.016 & 0.6 $\pm$ 0.019 \\
            A-CYPc[3] & 0.59 $\pm$ 0.003 & 0.62 $\pm$ 0.017 & 0.62 $\pm$ 0.016 & 0.55 $\pm$ 0.004 & 0.56 $\pm$ 0.002 & 0.57 $\pm$ 0.002 \\
            A-PM[1] & 0.44 $\pm$ 0.001 & 0.43 $\pm$ 0.001 & 0.44 $\pm$ 0.004 & 0.38 $\pm$ 0.002 & 0.38 $\pm$ 0.002 & 0.38 $\pm$ 0.002 \\
            A-PM[2] & 0.35 $\pm$ 0.002 & 0.33 $\pm$ 0.005 & 0.34 $\pm$ 0.006 & 0.28 $\pm$ 0.005 & 0.3 $\pm$ 0.006 & 0.3 $\pm$ 0.023 \\
            A-PM[3] & 0.34 $\pm$ 0.001 & 0.3 $\pm$ 0.003 & 0.3 $\pm$ 0.005 & 0.28 $\pm$ 0.006 & 0.27 $\pm$ 0.002 & 0.27 $\pm$ 0.002 \\
            A-SOL[1] & 0.68 $\pm$ 0.001 & 0.66 $\pm$ 0.006 & 0.66 $\pm$ 0.007 & 0.6 $\pm$ 0.002 & 0.59 $\pm$ 0.004 & 0.59 $\pm$ 0.004 \\
            A-SOL[2] & 0.68 $\pm$ 0.002 & 0.63 $\pm$ 0.007 & 0.63 $\pm$ 0.007 & 0.61 $\pm$ 0.03 & 0.6 $\pm$ 0.031 & 0.57 $\pm$ 0.02 \\
            A-SOL[3] & 0.67 $\pm$ 0.0 & 0.64 $\pm$ 0.006 & 0.64 $\pm$ 0.006 & 0.59 $\pm$ 0.027 & 0.58 $\pm$ 0.026 & 0.56 $\pm$ 0.013 \\
            A-hERG[1] & 0.61 $\pm$ 0.002 & 0.54 $\pm$ 0.003 & 0.54 $\pm$ 0.003 & 0.54 $\pm$ 0.024 & 0.52 $\pm$ 0.005 & 0.51 $\pm$ 0.004 \\
            A-hERG[2] & 0.59 $\pm$ 0.001 & 0.54 $\pm$ 0.004 & 0.54 $\pm$ 0.004 & 0.5 $\pm$ 0.002 & 0.5 $\pm$ 0.003 & 0.5 $\pm$ 0.001 \\
            A-hERG[3] & 0.54 $\pm$ 0.001 & 0.5 $\pm$ 0.003 & 0.51 $\pm$ 0.004 & 0.49 $\pm$ 0.011 & 0.48 $\pm$ 0.005 & 0.48 $\pm$ 0.004 \\
            A-logD[1] & 0.64 $\pm$ 0.004 & 0.62 $\pm$ 0.01 & 0.62 $\pm$ 0.01 & 0.59 $\pm$ 0.008 & 0.57 $\pm$ 0.004 & 0.57 $\pm$ 0.003 \\
            A-logD[2] & 0.67 $\pm$ 0.001 & 0.64 $\pm$ 0.003 & 0.64 $\pm$ 0.002 & 0.64 $\pm$ 0.031 & 0.62 $\pm$ 0.022 & 0.59 $\pm$ 0.016 \\
            A-logD[3] & 0.69 $\pm$ 0.001 & 0.66 $\pm$ 0.007 & 0.65 $\pm$ 0.008 & 0.53 $\pm$ 0.007 & 0.5 $\pm$ 0.004 & 0.5 $\pm$ 0.005 \\
            A-MS[1] & 0.64 $\pm$ 0.001 & 0.63 $\pm$ 0.004 & 0.63 $\pm$ 0.004 & 0.61 $\pm$ 0.018 & 0.6 $\pm$ 0.02 & 0.6 $\pm$ 0.028 \\
            A-MS[2] & 0.69 $\pm$ 0.002 & 0.66 $\pm$ 0.007 & 0.66 $\pm$ 0.006 & 0.68 $\pm$ 0.011 & 0.65 $\pm$ 0.015 & 0.65 $\pm$ 0.008 \\
            A-MS[3] & 0.64 $\pm$ 0.001 & 0.61 $\pm$ 0.004 & 0.61 $\pm$ 0.003 & 0.64 $\pm$ 0.014 & 0.64 $\pm$ 0.006 & 0.6 $\pm$ 0.005 \\
        \bottomrule
    \end{tabular}
\end{adjustbox}

\fontsize{9pt}{9pt}\selectfont
\begin{adjustbox}{rotate=90, caption = {\textbf{Summary of the BCE ($\downarrow$) scores for the uncertainty quantification models}. BCE results for all three temporal settings of all TB and ADME-T assays are reported. The temporal setting is indicated in brackets after the assay abbreviation.
Results for the uncalibrated uncertainty quantification models (deep ensembles (MLPE), MC dropout (MLPMC), and Bayesian neural network (BNN)), their Platt-scaled counterparts (MLPE-P, MLPMC-P and BNN-P) and the models calibrated with Venn-ABERS (VA) predictors (MLPE-VA, MLPMC-VA and BNN-VA) are reported. The models were trained with compounds from three time spans. Averages over 10 model repetitions are shown. 
} , float=table}
    \label{tab_supp:res_ue_bce}
    \centering
    \begin{tabular}{lrrrrrrrrr}
        \toprule
            Assay & MLPE & MLPE-P & MLPE-VA & MLPMC & MLPMC-P & MLPMC-VA & BNN & BNN-P & BNN-VA \\
        \midrule
            TB-1[1] & 0.7 $\pm$ 0.008 & 0.63 $\pm$ 0.021 & 0.56 $\pm$ 0.009 & 0.69 $\pm$ 0.014 & 0.65 $\pm$ 0.05 & 0.61 $\pm$ 0.065 & 0.7 $\pm$ 0.003 & 0.71 $\pm$ 0.001 & 0.78 $\pm$ 0.137 \\
            TB-1[2] & 0.44 $\pm$ 0.003 & 0.43 $\pm$ 0.001 & 0.44 $\pm$ 0.001 & 0.44 $\pm$ 0.013 & 0.43 $\pm$ 0.004 & 0.44 $\pm$ 0.004 & 0.74 $\pm$ 0.021 & 0.62 $\pm$ 0.004 & 0.62 $\pm$ 0.037 \\
            TB-1[3] & 0.67 $\pm$ 0.01 & 0.8 $\pm$ 0.004 & 0.77 $\pm$ 0.007 & 0.66 $\pm$ 0.023 & 0.79 $\pm$ 0.032 & 0.77 $\pm$ 0.038 & 0.69 $\pm$ 0.002 & 0.69 $\pm$ 0.001 & 0.75 $\pm$ 0.064 \\
            TB-2[1] & 0.64 $\pm$ 0.009 & 0.67 $\pm$ 0.006 & 0.64 $\pm$ 0.003 & 0.63 $\pm$ 0.011 & 0.65 $\pm$ 0.015 & 0.64 $\pm$ 0.007 & 0.69 $\pm$ 0.001 & 0.67 $\pm$ 0.0 & 0.72 $\pm$ 0.047 \\
            TB-2[2] & 0.54 $\pm$ 0.003 & 0.53 $\pm$ 0.004 & 0.53 $\pm$ 0.003 & 0.59 $\pm$ 0.053 & 0.53 $\pm$ 0.007 & 0.54 $\pm$ 0.009 & 0.68 $\pm$ 0.006 & 0.63 $\pm$ 0.002 & 0.7 $\pm$ 0.101 \\
            TB-2[3] & 0.63 $\pm$ 0.005 & 0.66 $\pm$ 0.003 & 0.7 $\pm$ 0.005 & 0.64 $\pm$ 0.024 & 0.67 $\pm$ 0.011 & 0.7 $\pm$ 0.013 & 0.72 $\pm$ 0.02 & 0.81 $\pm$ 0.007 & 0.81 $\pm$ 0.007 \\
            TB-3[1] & 0.5 $\pm$ 0.006 & 0.56 $\pm$ 0.001 & 0.55 $\pm$ 0.002 & 0.5 $\pm$ 0.014 & 0.55 $\pm$ 0.005 & 0.56 $\pm$ 0.008 & 0.53 $\pm$ 0.004 & 0.54 $\pm$ 0.004 & 0.54 $\pm$ 0.023 \\
            TB-3[2] & 0.45 $\pm$ 0.004 & 0.45 $\pm$ 0.003 & 0.46 $\pm$ 0.002 & 0.46 $\pm$ 0.015 & 0.45 $\pm$ 0.006 & 0.46 $\pm$ 0.005 & 0.43 $\pm$ 0.006 & 0.38 $\pm$ 0.008 & 0.35 $\pm$ 0.005 \\
            TB-3[3] & 0.19 $\pm$ 0.002 & 0.21 $\pm$ 0.001 & 0.21 $\pm$ 0.001 & 0.27 $\pm$ 0.012 & 0.2 $\pm$ 0.004 & 0.21 $\pm$ 0.003 & 0.28 $\pm$ 0.009 & 0.21 $\pm$ 0.011 & 0.21 $\pm$ 0.001 \\
            TB-4[1] & 0.3 $\pm$ 0.005 & 0.3 $\pm$ 0.002 & 0.3 $\pm$ 0.002 & 0.37 $\pm$ 0.013 & 0.31 $\pm$ 0.012 & 0.31 $\pm$ 0.009 & 0.4 $\pm$ 0.01 & 0.37 $\pm$ 0.006 & 0.32 $\pm$ 0.003 \\
            TB-4[2] & 0.45 $\pm$ 0.004 & 0.45 $\pm$ 0.005 & 0.46 $\pm$ 0.004 & 0.42 $\pm$ 0.018 & 0.44 $\pm$ 0.027 & 0.46 $\pm$ 0.018 & 0.43 $\pm$ 0.005 & 0.45 $\pm$ 0.006 & 0.52 $\pm$ 0.014 \\
            TB-4[3] & 0.25 $\pm$ 0.01 & 0.22 $\pm$ 0.004 & 0.22 $\pm$ 0.001 & 0.35 $\pm$ 0.058 & 0.23 $\pm$ 0.017 & 0.23 $\pm$ 0.013 & 0.33 $\pm$ 0.008 & 0.28 $\pm$ 0.003 & 0.22 $\pm$ 0.003 \\
            TB-5[1] & 0.52 $\pm$ 0.006 & 0.39 $\pm$ 0.001 & 0.37 $\pm$ 0.001 & 0.5 $\pm$ 0.036 & 0.39 $\pm$ 0.004 & 0.37 $\pm$ 0.003 & 0.46 $\pm$ 0.008 & 0.42 $\pm$ 0.002 & 0.43 $\pm$ 0.001 \\
            TB-5[2] & 0.57 $\pm$ 0.03 & 0.53 $\pm$ 0.003 & 0.52 $\pm$ 0.005 & 0.46 $\pm$ 0.022 & 0.52 $\pm$ 0.02 & 0.52 $\pm$ 0.022 & 0.49 $\pm$ 0.017 & 0.47 $\pm$ 0.012 & 0.51 $\pm$ 0.02 \\
            TB-5[3] & 0.3 $\pm$ 0.002 & 0.28 $\pm$ 0.001 & 0.29 $\pm$ 0.001 & 0.31 $\pm$ 0.007 & 0.28 $\pm$ 0.008 & 0.29 $\pm$ 0.006 & 0.38 $\pm$ 0.032 & 0.38 $\pm$ 0.024 & 0.35 $\pm$ 0.037 \\
            TB-6[1] & 0.54 $\pm$ 0.003 & 0.65 $\pm$ 0.006 & 0.66 $\pm$ 0.006 & 0.56 $\pm$ 0.008 & 0.65 $\pm$ 0.008 & 0.66 $\pm$ 0.008 & 0.58 $\pm$ 0.007 & 0.58 $\pm$ 0.002 & 0.58 $\pm$ 0.029 \\
            TB-6[2] & 0.45 $\pm$ 0.004 & 0.44 $\pm$ 0.004 & 0.45 $\pm$ 0.005 & 0.49 $\pm$ 0.025 & 0.44 $\pm$ 0.021 & 0.43 $\pm$ 0.027 & 0.45 $\pm$ 0.01 & 0.41 $\pm$ 0.014 & 0.39 $\pm$ 0.003 \\
            TB-6[3] & 0.29 $\pm$ 0.001 & 0.28 $\pm$ 0.0 & 0.29 $\pm$ 0.001 & 0.3 $\pm$ 0.006 & 0.28 $\pm$ 0.004 & 0.29 $\pm$ 0.003 & 0.37 $\pm$ 0.01 & 0.3 $\pm$ 0.002 & 0.31 $\pm$ 0.003 \\
            TB-7[1] & 0.66 $\pm$ 0.02 & 0.61 $\pm$ 0.026 & 0.58 $\pm$ 0.004 & 0.72 $\pm$ 0.032 & 0.72 $\pm$ 0.134 & 0.72 $\pm$ 0.125 & 0.69 $\pm$ 0.0 & 0.82 $\pm$ 0.001 & 0.83 $\pm$ 0.098 \\
            TB-7[2] & 0.46 $\pm$ 0.007 & 0.46 $\pm$ 0.005 & 0.42 $\pm$ 0.002 & 0.52 $\pm$ 0.027 & 0.47 $\pm$ 0.016 & 0.44 $\pm$ 0.012 & 0.6 $\pm$ 0.029 & 0.55 $\pm$ 0.019 & 0.46 $\pm$ 0.008 \\
            TB-7[3] & 0.3 $\pm$ 0.009 & 0.27 $\pm$ 0.002 & 0.27 $\pm$ 0.002 & 0.37 $\pm$ 0.048 & 0.27 $\pm$ 0.009 & 0.27 $\pm$ 0.008 & 0.69 $\pm$ 0.002 & 0.33 $\pm$ 0.002 & 0.36 $\pm$ 0.076 \\
        \midrule
            A-CYPa[1] & 0.54 $\pm$ 0.003 & 0.55 $\pm$ 0.004 & 0.56 $\pm$ 0.008 & 0.56 $\pm$ 0.012 & 0.55 $\pm$ 0.01 & 0.56 $\pm$ 0.016 & 0.6 $\pm$ 0.01 & 0.56 $\pm$ 0.002 & 0.57 $\pm$ 0.025 \\
            A-CYPa[2] & 0.48 $\pm$ 0.004 & 0.52 $\pm$ 0.023 & 0.58 $\pm$ 0.032 & 0.49 $\pm$ 0.01 & 0.48 $\pm$ 0.003 & 0.5 $\pm$ 0.004 & 0.47 $\pm$ 0.002 & 0.48 $\pm$ 0.002 & 0.5 $\pm$ 0.004 \\
            A-CYPa[3] & 0.52 $\pm$ 0.004 & 0.53 $\pm$ 0.001 & 0.52 $\pm$ 0.001 & 0.52 $\pm$ 0.012 & 0.52 $\pm$ 0.004 & 0.53 $\pm$ 0.004 & 0.52 $\pm$ 0.004 & 0.53 $\pm$ 0.006 & 0.53 $\pm$ 0.006 \\
            A-CYPb[1] & 0.49 $\pm$ 0.002 & 0.49 $\pm$ 0.0 & 0.49 $\pm$ 0.001 & 0.51 $\pm$ 0.008 & 0.5 $\pm$ 0.002 & 0.5 $\pm$ 0.003 & 0.51 $\pm$ 0.002 & 0.52 $\pm$ 0.001 & 0.51 $\pm$ 0.003 \\
            A-CYPb[2] & 0.48 $\pm$ 0.003 & 0.51 $\pm$ 0.002 & 0.51 $\pm$ 0.003 & 0.48 $\pm$ 0.015 & 0.52 $\pm$ 0.02 & 0.52 $\pm$ 0.025 & 0.46 $\pm$ 0.0 & 0.47 $\pm$ 0.003 & 0.49 $\pm$ 0.002 \\
            A-CYPb[3] & 0.38 $\pm$ 0.001 & 0.39 $\pm$ 0.0 & 0.39 $\pm$ 0.0 & 0.42 $\pm$ 0.018 & 0.39 $\pm$ 0.001 & 0.39 $\pm$ 0.002 & 0.39 $\pm$ 0.005 & 0.39 $\pm$ 0.002 & 0.38 $\pm$ 0.001 \\
            A-CYPc[1] & 0.59 $\pm$ 0.001 & 0.58 $\pm$ 0.001 & 0.58 $\pm$ 0.001 & 0.58 $\pm$ 0.006 & 0.58 $\pm$ 0.003 & 0.58 $\pm$ 0.002 & 0.63 $\pm$ 0.002 & 0.59 $\pm$ 0.0 & 0.59 $\pm$ 0.002 \\
            A-CYPc[2] & 0.6 $\pm$ 0.007 & 0.6 $\pm$ 0.004 & 0.58 $\pm$ 0.002 & 0.59 $\pm$ 0.01 & 0.62 $\pm$ 0.013 & 0.6 $\pm$ 0.017 & 0.62 $\pm$ 0.029 & 0.63 $\pm$ 0.032 & 0.62 $\pm$ 0.04 \\
            A-CYPc[3] & 0.55 $\pm$ 0.001 & 0.56 $\pm$ 0.0 & 0.56 $\pm$ 0.001 & 0.55 $\pm$ 0.002 & 0.56 $\pm$ 0.002 & 0.56 $\pm$ 0.002 & 0.54 $\pm$ 0.004 & 0.55 $\pm$ 0.004 & 0.55 $\pm$ 0.003 \\
            A-PM[1] & 0.38 $\pm$ 0.001 & 0.38 $\pm$ 0.0 & 0.38 $\pm$ 0.001 & 0.38 $\pm$ 0.003 & 0.37 $\pm$ 0.002 & 0.38 $\pm$ 0.002 & 0.42 $\pm$ 0.003 & 0.42 $\pm$ 0.002 & 0.37 $\pm$ 0.002 \\
            A-PM[2] & 0.28 $\pm$ 0.001 & 0.29 $\pm$ 0.001 & 0.29 $\pm$ 0.001 & 0.36 $\pm$ 0.018 & 0.29 $\pm$ 0.006 & 0.3 $\pm$ 0.005 & 0.3 $\pm$ 0.007 & 0.31 $\pm$ 0.012 & 0.29 $\pm$ 0.001 \\
            A-PM[3] & 0.27 $\pm$ 0.001 & 0.27 $\pm$ 0.0 & 0.27 $\pm$ 0.001 & 0.28 $\pm$ 0.006 & 0.27 $\pm$ 0.002 & 0.27 $\pm$ 0.002 & 0.29 $\pm$ 0.004 & 0.28 $\pm$ 0.003 & 0.26 $\pm$ 0.008 \\
            A-SOL[1] & 0.6 $\pm$ 0.001 & 0.59 $\pm$ 0.001 & 0.59 $\pm$ 0.001 & 0.59 $\pm$ 0.002 & 0.59 $\pm$ 0.004 & 0.59 $\pm$ 0.004 & 0.62 $\pm$ 0.039 & 0.6 $\pm$ 0.012 & 0.59 $\pm$ 0.013 \\
            A-SOL[2] & 0.56 $\pm$ 0.004 & 0.54 $\pm$ 0.005 & 0.54 $\pm$ 0.004 & 0.6 $\pm$ 0.013 & 0.59 $\pm$ 0.025 & 0.58 $\pm$ 0.023 & 0.69 $\pm$ 0.001 & 0.69 $\pm$ 0.001 & 0.72 $\pm$ 0.049 \\
            A-SOL[3] & 0.55 $\pm$ 0.004 & 0.55 $\pm$ 0.003 & 0.55 $\pm$ 0.003 & 0.57 $\pm$ 0.022 & 0.58 $\pm$ 0.022 & 0.56 $\pm$ 0.015 & 0.69 $\pm$ 0.001 & 0.7 $\pm$ 0.003 & 0.72 $\pm$ 0.068 \\
            A-hERG[1] & 0.52 $\pm$ 0.003 & 0.51 $\pm$ 0.001 & 0.51 $\pm$ 0.001 & 0.55 $\pm$ 0.022 & 0.52 $\pm$ 0.005 & 0.51 $\pm$ 0.005 & 0.63 $\pm$ 0.002 & 0.58 $\pm$ 0.0 & 0.57 $\pm$ 0.001 \\
            A-hERG[2] & 0.5 $\pm$ 0.001 & 0.49 $\pm$ 0.001 & 0.5 $\pm$ 0.0 & 0.51 $\pm$ 0.004 & 0.49 $\pm$ 0.002 & 0.5 $\pm$ 0.001 & 0.56 $\pm$ 0.017 & 0.54 $\pm$ 0.017 & 0.53 $\pm$ 0.014 \\
            A-hERG[3] & 0.47 $\pm$ 0.001 & 0.48 $\pm$ 0.001 & 0.47 $\pm$ 0.001 & 0.49 $\pm$ 0.01 & 0.48 $\pm$ 0.005 & 0.48 $\pm$ 0.004 & 0.51 $\pm$ 0.006 & 0.5 $\pm$ 0.004 & 0.5 $\pm$ 0.002 \\
            A-logD[1] & 0.58 $\pm$ 0.001 & 0.56 $\pm$ 0.001 & 0.57 $\pm$ 0.001 & 0.56 $\pm$ 0.003 & 0.56 $\pm$ 0.003 & 0.57 $\pm$ 0.003 & 0.61 $\pm$ 0.006 & 0.6 $\pm$ 0.004 & 0.6 $\pm$ 0.004 \\
            A-logD[2] & 0.6 $\pm$ 0.012 & 0.6 $\pm$ 0.009 & 0.57 $\pm$ 0.003 & 0.61 $\pm$ 0.022 & 0.6 $\pm$ 0.019 & 0.59 $\pm$ 0.009 & 0.68 $\pm$ 0.005 & 0.66 $\pm$ 0.008 & 0.66 $\pm$ 0.01 \\
            A-logD[3] & 0.52 $\pm$ 0.001 & 0.49 $\pm$ 0.003 & 0.49 $\pm$ 0.002 & 0.51 $\pm$ 0.005 & 0.5 $\pm$ 0.004 & 0.5 $\pm$ 0.004 & 0.51 $\pm$ 0.006 & 0.52 $\pm$ 0.004 & 0.5 $\pm$ 0.005 \\
            A-MS[1] & 0.6 $\pm$ 0.002 & 0.59 $\pm$ 0.001 & 0.59 $\pm$ 0.001 & 0.6 $\pm$ 0.018 & 0.6 $\pm$ 0.02 & 0.6 $\pm$ 0.028 & 0.61 $\pm$ 0.008 & 0.61 $\pm$ 0.004 & 0.6 $\pm$ 0.005 \\
            A-MS[2] & 0.67 $\pm$ 0.003 & 0.64 $\pm$ 0.002 & 0.64 $\pm$ 0.003 & 0.67 $\pm$ 0.009 & 0.65 $\pm$ 0.015 & 0.65 $\pm$ 0.015 & 0.66 $\pm$ 0.013 & 0.64 $\pm$ 0.007 & 0.64 $\pm$ 0.006 \\
            A-MS[3] & 0.62 $\pm$ 0.01 & 0.63 $\pm$ 0.008 & 0.61 $\pm$ 0.006 & 0.61 $\pm$ 0.01 & 0.61 $\pm$ 0.008 & 0.6 $\pm$ 0.005 & 0.58 $\pm$ 0.009 & 0.58 $\pm$ 0.007 & 0.58 $\pm$ 0.005 \\
        \bottomrule
        \end{tabular}
\end{adjustbox}

\fontsize{9pt}{9pt}\selectfont
\begin{adjustbox}{angle=0, caption = {\textbf{Summary of the ACE ($\downarrow$) scores for the baseline models}. ACE results for all three temporal settings of all TB and ADME-T assays are reported. The temporal setting is indicated in brackets after the assay abbreviation.
Results for the baselines (Random Forests (RF) and a multilayer perceptron (MLP)),  Platt scaled models (RF-P and MLP-P) and models calibrated with Venn-ABERS (VA) predictors (RF-VA and MLP-VA) are reported.
The models were trained with compounds from three time spans. Averages over 10 model repetitions are shown. 
} , float=table}
    \label{tab_supp:res_bl_ace}
    \centering
    \begin{tabular}{lrrrrrrr}
        \toprule
            Assay & RF & RF-P & RF-VA & MLP & MLP-P & MLP-VA \\
        \midrule
            TB-1[1] & 0.26 $\pm$ 0.011 & 0.16 $\pm$ 0.027 & 0.14 $\pm$ 0.018 & 0.23 $\pm$ 0.055 & 0.18 $\pm$ 0.039 & 0.12 $\pm$ 0.019 \\
            TB-1[2] & 0.27 $\pm$ 0.006 & 0.13 $\pm$ 0.011 & 0.13 $\pm$ 0.009 & 0.09 $\pm$ 0.016 & 0.08 $\pm$ 0.003 & 0.09 $\pm$ 0.008 \\
            TB-1[3] & 0.14 $\pm$ 0.018 & 0.21 $\pm$ 0.023 & 0.21 $\pm$ 0.006 & 0.16 $\pm$ 0.034 & 0.18 $\pm$ 0.028 & 0.19 $\pm$ 0.02 \\
            TB-2[1] & 0.09 $\pm$ 0.01 & 0.06 $\pm$ 0.014 & 0.09 $\pm$ 0.018 & 0.1 $\pm$ 0.016 & 0.11 $\pm$ 0.02 & 0.09 $\pm$ 0.008 \\
            TB-2[2] & 0.11 $\pm$ 0.011 & 0.06 $\pm$ 0.008 & 0.07 $\pm$ 0.019 & 0.14 $\pm$ 0.067 & 0.1 $\pm$ 0.012 & 0.09 $\pm$ 0.014 \\
            TB-2[3] & 0.23 $\pm$ 0.007 & 0.28 $\pm$ 0.011 & 0.27 $\pm$ 0.012 & 0.16 $\pm$ 0.025 & 0.2 $\pm$ 0.004 & 0.23 $\pm$ 0.006 \\
            TB-3[1] & 0.08 $\pm$ 0.014 & 0.23 $\pm$ 0.015 & 0.26 $\pm$ 0.01 & 0.08 $\pm$ 0.016 & 0.18 $\pm$ 0.005 & 0.17 $\pm$ 0.006 \\
            TB-3[2] & 0.2 $\pm$ 0.016 & 0.12 $\pm$ 0.014 & 0.11 $\pm$ 0.007 & 0.16 $\pm$ 0.018 & 0.15 $\pm$ 0.005 & 0.2 $\pm$ 0.061 \\
            TB-3[3] & 0.21 $\pm$ 0.009 & 0.07 $\pm$ 0.01 & 0.07 $\pm$ 0.01 & 0.06 $\pm$ 0.008 & 0.08 $\pm$ 0.007 & 0.09 $\pm$ 0.003 \\
            TB-4[1] & 0.24 $\pm$ 0.033 & 0.08 $\pm$ 0.018 & 0.08 $\pm$ 0.024 & 0.05 $\pm$ 0.016 & 0.05 $\pm$ 0.018 & 0.04 $\pm$ 0.007 \\
            TB-4[2] & 0.21 $\pm$ 0.026 & 0.09 $\pm$ 0.014 & 0.11 $\pm$ 0.015 & 0.11 $\pm$ 0.008 & 0.1 $\pm$ 0.012 & 0.1 $\pm$ 0.009 \\
            TB-4[3] & 0.25 $\pm$ 0.018 & 0.09 $\pm$ 0.015 & 0.06 $\pm$ 0.017 & 0.09 $\pm$ 0.066 & 0.07 $\pm$ 0.02 & 0.07 $\pm$ 0.032 \\
            TB-5[1] & 0.06 $\pm$ 0.009 & 0.16 $\pm$ 0.018 & 0.16 $\pm$ 0.025 & 0.09 $\pm$ 0.001 & 0.07 $\pm$ 0.001 & 0.07 $\pm$ 0.001 \\
            TB-5[2] & 0.07 $\pm$ 0.006 & 0.07 $\pm$ 0.013 & 0.08 $\pm$ 0.012 & 0.15 $\pm$ 0.012 & 0.12 $\pm$ 0.005 & 0.12 $\pm$ 0.005 \\
            TB-5[3] & 0.07 $\pm$ 0.01 & 0.06 $\pm$ 0.012 & 0.05 $\pm$ 0.009 & 0.05 $\pm$ 0.005 & 0.02 $\pm$ 0.003 & 0.03 $\pm$ 0.003 \\
            TB-6[1] & 0.08 $\pm$ 0.013 & 0.21 $\pm$ 0.019 & 0.23 $\pm$ 0.024 & 0.09 $\pm$ 0.011 & 0.23 $\pm$ 0.007 & 0.23 $\pm$ 0.006 \\
            TB-6[2] & 0.23 $\pm$ 0.008 & 0.15 $\pm$ 0.008 & 0.15 $\pm$ 0.019 & 0.17 $\pm$ 0.035 & 0.15 $\pm$ 0.013 & 0.19 $\pm$ 0.047 \\
            TB-6[3] & 0.2 $\pm$ 0.005 & 0.04 $\pm$ 0.01 & 0.04 $\pm$ 0.009 & 0.06 $\pm$ 0.013 & 0.04 $\pm$ 0.004 & 0.05 $\pm$ 0.004 \\
            TB-7[1] & 0.22 $\pm$ 0.006 & 0.16 $\pm$ 0.029 & 0.11 $\pm$ 0.035 & 0.16 $\pm$ 0.074 & 0.16 $\pm$ 0.107 & 0.17 $\pm$ 0.098 \\
            TB-7[2] & 0.21 $\pm$ 0.014 & 0.18 $\pm$ 0.016 & 0.09 $\pm$ 0.03 & 0.13 $\pm$ 0.045 & 0.12 $\pm$ 0.027 & 0.08 $\pm$ 0.02 \\
            TB-7[3] & 0.31 $\pm$ 0.009 & 0.11 $\pm$ 0.013 & 0.12 $\pm$ 0.026 & 0.11 $\pm$ 0.078 & 0.07 $\pm$ 0.012 & 0.07 $\pm$ 0.013 \\
        \midrule
            A-CYPa[1] & 0.09 $\pm$ 0.003 & 0.11 $\pm$ 0.011 & 0.09 $\pm$ 0.01 & 0.09 $\pm$ 0.014 & 0.1 $\pm$ 0.005 & 0.11 $\pm$ 0.027 \\
            A-CYPa[2] & 0.04 $\pm$ 0.013 & 0.06 $\pm$ 0.017 & 0.1 $\pm$ 0.029 & 0.06 $\pm$ 0.01 & 0.08 $\pm$ 0.004 & 0.08 $\pm$ 0.008 \\
            A-CYPa[3] & 0.08 $\pm$ 0.004 & 0.07 $\pm$ 0.009 & 0.06 $\pm$ 0.006 & 0.06 $\pm$ 0.012 & 0.07 $\pm$ 0.004 & 0.06 $\pm$ 0.005 \\
            A-CYPb[1] & 0.08 $\pm$ 0.005 & 0.07 $\pm$ 0.004 & 0.06 $\pm$ 0.006 & 0.06 $\pm$ 0.012 & 0.07 $\pm$ 0.003 & 0.07 $\pm$ 0.004 \\
            A-CYPb[2] & 0.04 $\pm$ 0.007 & 0.12 $\pm$ 0.011 & 0.14 $\pm$ 0.009 & 0.08 $\pm$ 0.016 & 0.11 $\pm$ 0.008 & 0.11 $\pm$ 0.01 \\
            A-CYPb[3] & 0.05 $\pm$ 0.005 & 0.05 $\pm$ 0.006 & 0.05 $\pm$ 0.006 & 0.05 $\pm$ 0.017 & 0.05 $\pm$ 0.003 & 0.05 $\pm$ 0.004 \\
            A-CYPc[1] & 0.12 $\pm$ 0.003 & 0.06 $\pm$ 0.013 & 0.07 $\pm$ 0.015 & 0.08 $\pm$ 0.003 & 0.07 $\pm$ 0.003 & 0.07 $\pm$ 0.001 \\
            A-CYPc[2] & 0.11 $\pm$ 0.012 & 0.1 $\pm$ 0.008 & 0.08 $\pm$ 0.007 & 0.12 $\pm$ 0.021 & 0.1 $\pm$ 0.005 & 0.09 $\pm$ 0.006 \\
            A-CYPc[3] & 0.07 $\pm$ 0.013 & 0.13 $\pm$ 0.011 & 0.13 $\pm$ 0.012 & 0.05 $\pm$ 0.012 & 0.09 $\pm$ 0.002 & 0.09 $\pm$ 0.002 \\
            A-PM[1] & 0.05 $\pm$ 0.003 & 0.04 $\pm$ 0.004 & 0.05 $\pm$ 0.003 & 0.02 $\pm$ 0.005 & 0.02 $\pm$ 0.001 & 0.02 $\pm$ 0.003 \\
            A-PM[2] & 0.1 $\pm$ 0.003 & 0.06 $\pm$ 0.005 & 0.07 $\pm$ 0.006 & 0.04 $\pm$ 0.007 & 0.07 $\pm$ 0.007 & 0.08 $\pm$ 0.024 \\
            A-PM[3] & 0.09 $\pm$ 0.001 & 0.02 $\pm$ 0.005 & 0.03 $\pm$ 0.008 & 0.04 $\pm$ 0.01 & 0.03 $\pm$ 0.002 & 0.03 $\pm$ 0.002 \\
            A-SOL[1] & 0.1 $\pm$ 0.009 & 0.05 $\pm$ 0.009 & 0.03 $\pm$ 0.005 & 0.06 $\pm$ 0.004 & 0.03 $\pm$ 0.003 & 0.02 $\pm$ 0.003 \\
            A-SOL[2] & 0.13 $\pm$ 0.006 & 0.04 $\pm$ 0.007 & 0.04 $\pm$ 0.007 & 0.09 $\pm$ 0.021 & 0.09 $\pm$ 0.027 & 0.04 $\pm$ 0.007 \\
            A-SOL[3] & 0.13 $\pm$ 0.008 & 0.07 $\pm$ 0.005 & 0.07 $\pm$ 0.006 & 0.08 $\pm$ 0.019 & 0.06 $\pm$ 0.023 & 0.04 $\pm$ 0.005 \\
            A-hERG[1] & 0.17 $\pm$ 0.005 & 0.04 $\pm$ 0.004 & 0.05 $\pm$ 0.006 & 0.09 $\pm$ 0.031 & 0.03 $\pm$ 0.007 & 0.02 $\pm$ 0.003 \\
            A-hERG[2] & 0.11 $\pm$ 0.004 & 0.04 $\pm$ 0.008 & 0.04 $\pm$ 0.01 & 0.05 $\pm$ 0.004 & 0.03 $\pm$ 0.009 & 0.02 $\pm$ 0.003 \\
            A-hERG[3] & 0.1 $\pm$ 0.001 & 0.03 $\pm$ 0.003 & 0.05 $\pm$ 0.006 & 0.07 $\pm$ 0.01 & 0.06 $\pm$ 0.007 & 0.05 $\pm$ 0.007 \\
            A-logD[1] & 0.08 $\pm$ 0.013 & 0.03 $\pm$ 0.008 & 0.03 $\pm$ 0.008 & 0.08 $\pm$ 0.008 & 0.05 $\pm$ 0.002 & 0.04 $\pm$ 0.002 \\
            A-logD[2] & 0.09 $\pm$ 0.007 & 0.05 $\pm$ 0.006 & 0.05 $\pm$ 0.007 & 0.1 $\pm$ 0.021 & 0.08 $\pm$ 0.02 & 0.06 $\pm$ 0.016 \\
            A-logD[3] & 0.12 $\pm$ 0.012 & 0.08 $\pm$ 0.01 & 0.05 $\pm$ 0.005 & 0.07 $\pm$ 0.007 & 0.03 $\pm$ 0.003 & 0.03 $\pm$ 0.003 \\
            A-MS[1] & 0.07 $\pm$ 0.011 & 0.03 $\pm$ 0.006 & 0.03 $\pm$ 0.009 & 0.08 $\pm$ 0.022 & 0.04 $\pm$ 0.006 & 0.05 $\pm$ 0.02 \\
            A-MS[2] & 0.11 $\pm$ 0.007 & 0.05 $\pm$ 0.009 & 0.04 $\pm$ 0.004 & 0.12 $\pm$ 0.017 & 0.06 $\pm$ 0.017 & 0.06 $\pm$ 0.007 \\
            A-MS[3] & 0.11 $\pm$ 0.006 & 0.06 $\pm$ 0.006 & 0.04 $\pm$ 0.003 & 0.08 $\pm$ 0.01 & 0.08 $\pm$ 0.009 & 0.05 $\pm$ 0.005 \\
        \bottomrule
        \end{tabular}
\end{adjustbox}

\fontsize{9pt}{9pt}\selectfont
\begin{adjustbox}{angle=90, caption = {\textbf{Summary of the ACE ($\downarrow$) scores for the uncertainty quantification models}. ACE results for all three temporal settings of all TB and ADME-T assays are reported. The temporal setting is indicated in brackets after the assay abbreviation.
Results for the uncalibrated uncertainty quantification models (deep ensembles (MLPE), MC dropout (MLPMC), and Bayesian neural network (BNN)), their Platt-scaled counterparts (MLPE-P, MLPMC-P and BNN-P) and the models calibrated with Venn-ABERS (VA) predictors (MLPE-VA, MLPMC-VA and BNN-VA) are reported. The models were trained with compounds from three time spans. Averages over 10 model repetitions are shown. 
} , float=table}
    \label{tab_supp:res_ue_ace}
    \centering
    \begin{tabular}{lrrrrrrrrr}
        \toprule
            Assay & MLPE & MLPE-P & MLPE-VA & MLPMC & MLPMC-P & MLPMC-VA & BNN & BNN-P & BNN-VA \\
        \midrule
            TB-1[1] & 0.28 $\pm$ 0.006 & 0.21 $\pm$ 0.021 & 0.13 $\pm$ 0.015 & 0.22 $\pm$ 0.055 & 0.17 $\pm$ 0.039 & 0.13 $\pm$ 0.022 & 0.09 $\pm$ 0.007 & 0.11 $\pm$ 0.004 & 0.18 $\pm$ 0.105 \\
            TB-1[2] & 0.1 $\pm$ 0.004 & 0.08 $\pm$ 0.001 & 0.09 $\pm$ 0.005 & 0.1 $\pm$ 0.019 & 0.08 $\pm$ 0.003 & 0.09 $\pm$ 0.008 & 0.31 $\pm$ 0.021 & 0.21 $\pm$ 0.014 & 0.2 $\pm$ 0.031 \\
            TB-1[3] & 0.13 $\pm$ 0.013 & 0.21 $\pm$ 0.002 & 0.18 $\pm$ 0.004 & 0.1 $\pm$ 0.022 & 0.2 $\pm$ 0.019 & 0.19 $\pm$ 0.018 & 0.15 $\pm$ 0.003 & 0.12 $\pm$ 0.002 & 0.21 $\pm$ 0.034 \\
            TB-2[1] & 0.09 $\pm$ 0.008 & 0.1 $\pm$ 0.004 & 0.09 $\pm$ 0.004 & 0.08 $\pm$ 0.012 & 0.1 $\pm$ 0.01 & 0.08 $\pm$ 0.01 & 0.11 $\pm$ 0.003 & 0.04 $\pm$ 0.011 & 0.17 $\pm$ 0.05 \\
            TB-2[2] & 0.11 $\pm$ 0.006 & 0.09 $\pm$ 0.006 & 0.08 $\pm$ 0.006 & 0.18 $\pm$ 0.051 & 0.09 $\pm$ 0.007 & 0.08 $\pm$ 0.01 & 0.18 $\pm$ 0.009 & 0.09 $\pm$ 0.004 & 0.2 $\pm$ 0.101 \\
            TB-2[3] & 0.17 $\pm$ 0.005 & 0.2 $\pm$ 0.001 & 0.23 $\pm$ 0.002 & 0.18 $\pm$ 0.023 & 0.21 $\pm$ 0.004 & 0.23 $\pm$ 0.005 & 0.2 $\pm$ 0.022 & 0.28 $\pm$ 0.007 & 0.24 $\pm$ 0.007 \\
            TB-3[1] & 0.09 $\pm$ 0.009 & 0.18 $\pm$ 0.004 & 0.17 $\pm$ 0.002 & 0.1 $\pm$ 0.022 & 0.17 $\pm$ 0.007 & 0.17 $\pm$ 0.007 & 0.08 $\pm$ 0.012 & 0.1 $\pm$ 0.007 & 0.09 $\pm$ 0.04 \\
            TB-3[2] & 0.15 $\pm$ 0.005 & 0.15 $\pm$ 0.002 & 0.16 $\pm$ 0.002 & 0.18 $\pm$ 0.016 & 0.15 $\pm$ 0.006 & 0.16 $\pm$ 0.008 & 0.19 $\pm$ 0.006 & 0.14 $\pm$ 0.013 & 0.1 $\pm$ 0.007 \\
            TB-3[3] & 0.06 $\pm$ 0.002 & 0.08 $\pm$ 0.001 & 0.09 $\pm$ 0.001 & 0.16 $\pm$ 0.011 & 0.07 $\pm$ 0.004 & 0.08 $\pm$ 0.003 & 0.16 $\pm$ 0.009 & 0.09 $\pm$ 0.012 & 0.09 $\pm$ 0.002 \\
            TB-4[1] & 0.04 $\pm$ 0.007 & 0.03 $\pm$ 0.003 & 0.04 $\pm$ 0.003 & 0.11 $\pm$ 0.012 & 0.03 $\pm$ 0.012 & 0.04 $\pm$ 0.006 & 0.16 $\pm$ 0.009 & 0.12 $\pm$ 0.005 & 0.07 $\pm$ 0.003 \\
            TB-4[2] & 0.1 $\pm$ 0.003 & 0.1 $\pm$ 0.002 & 0.1 $\pm$ 0.003 & 0.07 $\pm$ 0.006 & 0.09 $\pm$ 0.015 & 0.11 $\pm$ 0.011 & 0.06 $\pm$ 0.005 & 0.09 $\pm$ 0.012 & 0.14 $\pm$ 0.006 \\
            TB-4[3] & 0.08 $\pm$ 0.012 & 0.05 $\pm$ 0.004 & 0.05 $\pm$ 0.003 & 0.18 $\pm$ 0.044 & 0.06 $\pm$ 0.009 & 0.05 $\pm$ 0.005 & 0.15 $\pm$ 0.007 & 0.1 $\pm$ 0.005 & 0.05 $\pm$ 0.003 \\
            TB-5[1] & 0.09 $\pm$ 0.0 & 0.07 $\pm$ 0.0 & 0.07 $\pm$ 0.0 & 0.09 $\pm$ 0.002 & 0.07 $\pm$ 0.001 & 0.07 $\pm$ 0.002 & 0.18 $\pm$ 0.009 & 0.13 $\pm$ 0.004 & 0.13 $\pm$ 0.004 \\
            TB-5[2] & 0.14 $\pm$ 0.009 & 0.12 $\pm$ 0.001 & 0.12 $\pm$ 0.002 & 0.08 $\pm$ 0.024 & 0.12 $\pm$ 0.006 & 0.12 $\pm$ 0.005 & 0.1 $\pm$ 0.016 & 0.08 $\pm$ 0.007 & 0.1 $\pm$ 0.014 \\
            TB-5[3] & 0.05 $\pm$ 0.002 & 0.02 $\pm$ 0.001 & 0.03 $\pm$ 0.002 & 0.05 $\pm$ 0.005 & 0.02 $\pm$ 0.002 & 0.03 $\pm$ 0.003 & 0.08 $\pm$ 0.015 & 0.09 $\pm$ 0.011 & 0.03 $\pm$ 0.016 \\
            TB-6[1] & 0.1 $\pm$ 0.005 & 0.23 $\pm$ 0.004 & 0.24 $\pm$ 0.006 & 0.12 $\pm$ 0.015 & 0.22 $\pm$ 0.01 & 0.23 $\pm$ 0.009 & 0.09 $\pm$ 0.018 & 0.09 $\pm$ 0.006 & 0.08 $\pm$ 0.053 \\
            TB-6[2] & 0.16 $\pm$ 0.004 & 0.14 $\pm$ 0.003 & 0.15 $\pm$ 0.004 & 0.2 $\pm$ 0.023 & 0.14 $\pm$ 0.017 & 0.14 $\pm$ 0.024 & 0.2 $\pm$ 0.01 & 0.16 $\pm$ 0.017 & 0.11 $\pm$ 0.002 \\
            TB-6[3] & 0.05 $\pm$ 0.001 & 0.04 $\pm$ 0.001 & 0.05 $\pm$ 0.001 & 0.07 $\pm$ 0.012 & 0.04 $\pm$ 0.004 & 0.05 $\pm$ 0.003 & 0.15 $\pm$ 0.011 & 0.04 $\pm$ 0.005 & 0.06 $\pm$ 0.006 \\
            TB-7[1] & 0.19 $\pm$ 0.023 & 0.12 $\pm$ 0.03 & 0.05 $\pm$ 0.015 & 0.14 $\pm$ 0.08 & 0.16 $\pm$ 0.105 & 0.17 $\pm$ 0.09 & 0.03 $\pm$ 0.003 & 0.24 $\pm$ 0.001 & 0.25 $\pm$ 0.063 \\
            TB-7[2] & 0.12 $\pm$ 0.008 & 0.12 $\pm$ 0.006 & 0.06 $\pm$ 0.006 & 0.18 $\pm$ 0.023 & 0.12 $\pm$ 0.013 & 0.07 $\pm$ 0.017 & 0.25 $\pm$ 0.024 & 0.21 $\pm$ 0.015 & 0.08 $\pm$ 0.011 \\
            TB-7[3] & 0.11 $\pm$ 0.01 & 0.07 $\pm$ 0.002 & 0.07 $\pm$ 0.004 & 0.18 $\pm$ 0.041 & 0.06 $\pm$ 0.009 & 0.06 $\pm$ 0.011 & 0.42 $\pm$ 0.001 & 0.11 $\pm$ 0.003 & 0.14 $\pm$ 0.067 \\
            \midrule
            A-CYPa[1] & 0.08 $\pm$ 0.003 & 0.1 $\pm$ 0.005 & 0.09 $\pm$ 0.01 & 0.11 $\pm$ 0.025 & 0.1 $\pm$ 0.009 & 0.09 $\pm$ 0.017 & 0.18 $\pm$ 0.013 & 0.11 $\pm$ 0.007 & 0.11 $\pm$ 0.023 \\
            A-CYPa[2] & 0.08 $\pm$ 0.011 & 0.12 $\pm$ 0.019 & 0.14 $\pm$ 0.02 & 0.09 $\pm$ 0.022 & 0.08 $\pm$ 0.006 & 0.08 $\pm$ 0.006 & 0.04 $\pm$ 0.01 & 0.07 $\pm$ 0.006 & 0.11 $\pm$ 0.006 \\
            A-CYPa[3] & 0.05 $\pm$ 0.001 & 0.07 $\pm$ 0.002 & 0.07 $\pm$ 0.003 & 0.07 $\pm$ 0.025 & 0.06 $\pm$ 0.002 & 0.06 $\pm$ 0.004 & 0.04 $\pm$ 0.008 & 0.06 $\pm$ 0.009 & 0.06 $\pm$ 0.002 \\
            A-CYPb[1] & 0.06 $\pm$ 0.003 & 0.07 $\pm$ 0.001 & 0.07 $\pm$ 0.002 & 0.09 $\pm$ 0.019 & 0.06 $\pm$ 0.003 & 0.07 $\pm$ 0.003 & 0.05 $\pm$ 0.005 & 0.06 $\pm$ 0.004 & 0.05 $\pm$ 0.002 \\
            A-CYPb[2] & 0.06 $\pm$ 0.003 & 0.11 $\pm$ 0.002 & 0.11 $\pm$ 0.003 & 0.08 $\pm$ 0.021 & 0.11 $\pm$ 0.01 & 0.12 $\pm$ 0.012 & 0.04 $\pm$ 0.004 & 0.06 $\pm$ 0.01 & 0.09 $\pm$ 0.002 \\
            A-CYPb[3] & 0.04 $\pm$ 0.003 & 0.05 $\pm$ 0.001 & 0.05 $\pm$ 0.003 & 0.09 $\pm$ 0.032 & 0.05 $\pm$ 0.002 & 0.05 $\pm$ 0.004 & 0.05 $\pm$ 0.011 & 0.05 $\pm$ 0.006 & 0.04 $\pm$ 0.003 \\
            A-CYPc[1] & 0.08 $\pm$ 0.001 & 0.07 $\pm$ 0.001 & 0.07 $\pm$ 0.001 & 0.06 $\pm$ 0.003 & 0.07 $\pm$ 0.003 & 0.07 $\pm$ 0.001 & 0.17 $\pm$ 0.003 & 0.1 $\pm$ 0.004 & 0.07 $\pm$ 0.007 \\
            A-CYPc[2] & 0.12 $\pm$ 0.006 & 0.11 $\pm$ 0.003 & 0.09 $\pm$ 0.001 & 0.09 $\pm$ 0.016 & 0.11 $\pm$ 0.008 & 0.09 $\pm$ 0.005 & 0.07 $\pm$ 0.007 & 0.09 $\pm$ 0.005 & 0.08 $\pm$ 0.012 \\
            A-CYPc[3] & 0.05 $\pm$ 0.002 & 0.08 $\pm$ 0.001 & 0.09 $\pm$ 0.002 & 0.05 $\pm$ 0.008 & 0.08 $\pm$ 0.002 & 0.09 $\pm$ 0.003 & 0.06 $\pm$ 0.01 & 0.08 $\pm$ 0.01 & 0.08 $\pm$ 0.004 \\
            A-PM[1] & 0.02 $\pm$ 0.002 & 0.01 $\pm$ 0.001 & 0.02 $\pm$ 0.002 & 0.03 $\pm$ 0.006 & 0.01 $\pm$ 0.001 & 0.02 $\pm$ 0.002 & 0.13 $\pm$ 0.004 & 0.13 $\pm$ 0.004 & 0.06 $\pm$ 0.006 \\
            A-PM[2] & 0.05 $\pm$ 0.003 & 0.07 $\pm$ 0.001 & 0.07 $\pm$ 0.001 & 0.13 $\pm$ 0.02 & 0.06 $\pm$ 0.007 & 0.07 $\pm$ 0.007 & 0.07 $\pm$ 0.01 & 0.08 $\pm$ 0.015 & 0.07 $\pm$ 0.001 \\
            A-PM[3] & 0.04 $\pm$ 0.002 & 0.03 $\pm$ 0.001 & 0.03 $\pm$ 0.001 & 0.05 $\pm$ 0.009 & 0.03 $\pm$ 0.002 & 0.03 $\pm$ 0.002 & 0.07 $\pm$ 0.007 & 0.04 $\pm$ 0.005 & 0.02 $\pm$ 0.003 \\
            A-SOL[1] & 0.05 $\pm$ 0.001 & 0.04 $\pm$ 0.001 & 0.02 $\pm$ 0.003 & 0.05 $\pm$ 0.004 & 0.04 $\pm$ 0.004 & 0.02 $\pm$ 0.004 & 0.09 $\pm$ 0.047 & 0.04 $\pm$ 0.008 & 0.02 $\pm$ 0.003 \\
            A-SOL[2] & 0.08 $\pm$ 0.007 & 0.04 $\pm$ 0.009 & 0.03 $\pm$ 0.004 & 0.09 $\pm$ 0.016 & 0.07 $\pm$ 0.012 & 0.04 $\pm$ 0.01 & 0.04 $\pm$ 0.005 & 0.03 $\pm$ 0.007 & 0.08 $\pm$ 0.068 \\
            A-SOL[3] & 0.04 $\pm$ 0.006 & 0.05 $\pm$ 0.003 & 0.03 $\pm$ 0.002 & 0.06 $\pm$ 0.018 & 0.07 $\pm$ 0.017 & 0.04 $\pm$ 0.004 & 0.05 $\pm$ 0.003 & 0.1 $\pm$ 0.008 & 0.11 $\pm$ 0.066 \\
            A-hERG[1] & 0.06 $\pm$ 0.009 & 0.03 $\pm$ 0.001 & 0.02 $\pm$ 0.002 & 0.11 $\pm$ 0.031 & 0.04 $\pm$ 0.006 & 0.02 $\pm$ 0.003 & 0.16 $\pm$ 0.003 & 0.05 $\pm$ 0.002 & 0.03 $\pm$ 0.006 \\
            A-hERG[2] & 0.04 $\pm$ 0.001 & 0.03 $\pm$ 0.003 & 0.02 $\pm$ 0.002 & 0.06 $\pm$ 0.008 & 0.03 $\pm$ 0.003 & 0.02 $\pm$ 0.003 & 0.08 $\pm$ 0.005 & 0.04 $\pm$ 0.013 & 0.02 $\pm$ 0.005 \\
            A-hERG[3] & 0.05 $\pm$ 0.003 & 0.06 $\pm$ 0.001 & 0.05 $\pm$ 0.001 & 0.09 $\pm$ 0.016 & 0.06 $\pm$ 0.005 & 0.05 $\pm$ 0.008 & 0.06 $\pm$ 0.014 & 0.03 $\pm$ 0.014 & 0.03 $\pm$ 0.002 \\
            A-logD[1] & 0.08 $\pm$ 0.001 & 0.05 $\pm$ 0.001 & 0.04 $\pm$ 0.001 & 0.05 $\pm$ 0.006 & 0.05 $\pm$ 0.002 & 0.04 $\pm$ 0.003 & 0.07 $\pm$ 0.002 & 0.05 $\pm$ 0.01 & 0.04 $\pm$ 0.003 \\
            A-logD[2] & 0.08 $\pm$ 0.011 & 0.08 $\pm$ 0.009 & 0.05 $\pm$ 0.002 & 0.08 $\pm$ 0.018 & 0.08 $\pm$ 0.016 & 0.06 $\pm$ 0.006 & 0.11 $\pm$ 0.01 & 0.08 $\pm$ 0.005 & 0.06 $\pm$ 0.003 \\
            A-logD[3] & 0.07 $\pm$ 0.002 & 0.03 $\pm$ 0.001 & 0.03 $\pm$ 0.001 & 0.05 $\pm$ 0.007 & 0.03 $\pm$ 0.003 & 0.03 $\pm$ 0.003 & 0.03 $\pm$ 0.011 & 0.06 $\pm$ 0.008 & 0.02 $\pm$ 0.003 \\
            A-MS[1] & 0.08 $\pm$ 0.005 & 0.04 $\pm$ 0.001 & 0.05 $\pm$ 0.002 & 0.07 $\pm$ 0.014 & 0.04 $\pm$ 0.007 & 0.05 $\pm$ 0.02 & 0.06 $\pm$ 0.011 & 0.07 $\pm$ 0.008 & 0.04 $\pm$ 0.003 \\
            A-MS[2] & 0.12 $\pm$ 0.003 & 0.05 $\pm$ 0.002 & 0.06 $\pm$ 0.002 & 0.1 $\pm$ 0.016 & 0.06 $\pm$ 0.007 & 0.06 $\pm$ 0.005 & 0.11 $\pm$ 0.016 & 0.06 $\pm$ 0.017 & 0.06 $\pm$ 0.001 \\
            A-MS[3] & 0.07 $\pm$ 0.006 & 0.08 $\pm$ 0.006 & 0.05 $\pm$ 0.002 & 0.06 $\pm$ 0.012 & 0.07 $\pm$ 0.009 & 0.05 $\pm$ 0.005 & 0.04 $\pm$ 0.013 & 0.04 $\pm$ 0.009 & 0.04 $\pm$ 0.003 \\
        \bottomrule
    \end{tabular}
\end{adjustbox}

\end{document}